\documentclass[10pt,twocolumn,letterpaper]{article}

\usepackage[pagenumbers]{wacv} 

\newcommand{\red}[1]{#1}

\definecolor{unverified}{rgb}{0.05,0.25,0.85}

\usepackage{pifont}
\newcommand{\cmark}{\textcolor{green!60!black}{\ding{51}}} %
\newcommand{\xmark}{\textcolor{red!80!black}{\ding{55}}}   %

\usepackage{colortbl} %

\newcommand{\claro}{\mbox{CLARO}}

\usepackage{booktabs}
\usepackage{multirow}
\usepackage[numbers]{natbib}
\usepackage{tikz}
\usepackage{enumitem}
\usepackage{pgfplots}
\pgfplotsset{compat=1.17}
\usetikzlibrary{arrows.meta,positioning,shapes.geometric,backgrounds,fit,calc}

\newcommand{\best}[1]{\textbf{#1}}
\newcommand{\nd}[1]{\underline{#1}}

\definecolor{wacvblue}{rgb}{0.21,0.49,0.74}
\usepackage[pagebackref,breaklinks,colorlinks,allcolors=wacvblue]{hyperref}

\def\wacvPaperID{1850} 
\def\confName{WACV}
\def\confYear{2027}

\title{CLARO: Clutter Aware Robot Orchestrator for Open-Vocabulary \\ Grasping using VLM Agents}

\author{Sungsu Kim$^{1,*}$\hspace{0.65em} Vineet Bhat$^{2,*}$\hspace{0.65em} Prashanth Krishnamurthy$^{2}$\hspace{0.65em} Ramesh Karri$^{2}$\hspace{0.65em} Farshad Khorrami$^{2}$%
\\[0.6em]
$^{1}$New York University Abu Dhabi, Abu Dhabi, UAE \\
$^{2}$New York University Tandon School of Engineering \\
{\normalsize Dept. of Electrical and Computer Engineering, 5 MetroTech Center, Brooklyn, NY 11201, USA} \\
{\tt\small \{sk9146, vrb9107, prashanth.krishnamurthy, rkarri, khorrami\}@nyu.edu}
}

\begin{document}
\maketitle

{\renewcommand{\thefootnote}{}\footnotetext{$^{*}$Equal contribution.}}

\begin{abstract}
Robots that follow natural-language instructions unlock a wide range of
autonomous applications in households, warehouses, and labs. Real-world
environments are diverse and often heavily cluttered, requiring robots to plan over multiple steps, act, and replan as they work toward a goal.
We present \claro{} (\textbf{CL}utter-\textbf{A}ware \textbf{R}obot
\textbf{O}rchestrator), a VLM agent that orchestrates a set of perception and
manipulation skills: open-vocabulary object detection, segmentation, $6$-DoF grasping,
and non-prehensile pushing to locate and retrieve a language-specified target in clutter. \claro{} maintains a memory of its past actions to support long-horizon execution, and emits manipulation
targets (grasp poses/3D waypoints) allowing transfer across robot arms without task-specific training. We evaluate \claro{} against a range of methods
on a dynamic simulation environment and standard image-based benchmarks, demonstrating
strong generalization across object classes, lighting, and clutter levels.
\claro{} attains an average success rate of $89.1\%$, a relative
improvement of over $16\%$ on the strongest competing method in high cluttered scenes. We further analyze performance along grounding and grasping axes to
characterize where each method succeeds and fails. Finally, we deploy \claro{} on a
robot arm across diverse cluttered scenes, achieving a success rate of $88.8\%$, and demonstrating strong sim2real generalizability.
\end{abstract}

\vspace{-20pt}

\section{Introduction}
\label{sec:intro}

Grasping objects specified by natural-language is a
long-standing goal of robotics and embodied AI~\cite{danielczuk2019mechanical}.
An instruction like ``hand me the small screwdriver behind the mug'' requires
grounding an open-vocabulary referred object and reasoning about its spatial
position in the scene. Vision Language Models (VLMs) have brought this
goal within reach: by jointly reasoning over images and language, they form a
perception backbone that grounds free-form text instructions to the robot's visual
observations~\cite{Clark_2026_CVPR,team2025gemini}. VLMs have been adapted for Language-Driven Robot Manipulation (LDRM) in tasks such as pick-and-place and articulated-object
manipulation~\cite{Bhat_2026_WACV,intelligence2025pi_}.
\begin{figure}[ht]
  \centering
  \setlength{\tabcolsep}{2pt}
  \renewcommand{\arraystretch}{0.6}
  \includegraphics[width=\linewidth]
  {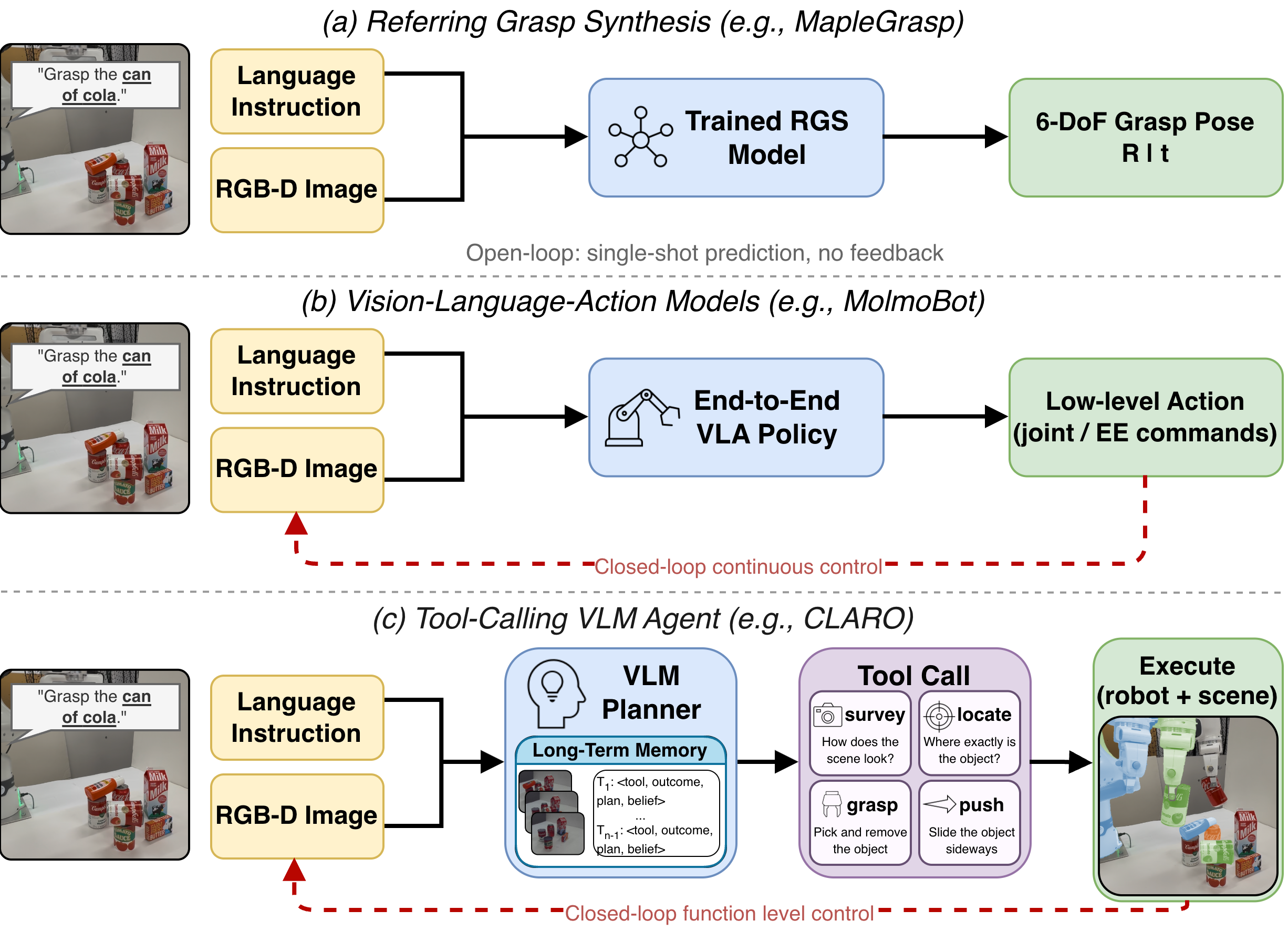}
  \caption{\textbf{Approaches for Language-Driven Robot Manipulation (LDRM) are distinguished by whether and how they close the loop with the environment.} \emph{(a) Referring grasp synthesis (RGS)} predicts a grasp pose in a single open-loop pass. \emph{(b)
Vision language action policies} close the loop at the action level, mapping
observations and language to low-level motor commands. \emph{(c) Tool-calling
VLM agents} such as \claro{} close it at the tool level with a VLM planner invoking perception and manipulation tools
conditioned on a past memory of tool calls and outcomes.}
  \label{fig:ldg-examples}
\end{figure}

\begin{table*}[t]
  \centering
  \footnotesize
  \setlength{\tabcolsep}{6pt}
  \renewcommand{\arraystretch}{1.3}
  \begin{tabular}{l@{\hspace{6pt}}|@{\hspace{6pt}}c ccccc}
    \toprule
    \textbf{Method} & \textbf{\shortstack{Grasp\\Modality}}
      & \textbf{\shortstack{Open-\\Vocab.}}
      & \textbf{\shortstack{Closed-Loop\\Planning}}
      & \textbf{\shortstack{Long-Term\\Memory}}
      & \textbf{\shortstack{Non-Prehensile\\Manip.}}
      & \textbf{\shortstack{Cross-Arm\\Deploy.}} \\
    \midrule
    \multicolumn{7}{c}{\emph{Referring Grasp Synthesis methods}}\\
    CROG~\cite{tziafas2023language}          & 4-DoF & \xmark & \xmark & \xmark & \xmark & \cmark \\
    MapleGrasp~\cite{Bhat_2026_WACV}    & 6-DoF & \xmark & \xmark & \xmark & \xmark & \cmark \\
    \red{UnoGrasp~\cite{jiao2025obstruction}} & \red{6-DoF} & \xmark & \cmark & \xmark & \xmark & \cmark \\
    \midrule

    \multicolumn{7}{c}{\red{\emph{Language-agnostic grasp generation}}}\\
    \red{GraspGen~\cite{murali2025graspgen}} & \red{6-DoF} & \xmark & \xmark & \xmark & \xmark & \cmark \\
    \midrule

    \multicolumn{7}{c}{\emph{Vision--Language--Action policies}}\\
    $\pi_{0.5}$~\cite{intelligence2025pi_}& 6-DoF & \xmark & \cmark & \xmark & \cmark & \xmark \\
    MolmoBot~\cite{deshpande2026molmob0t}                 & 6-DoF & \xmark & \cmark & \cmark & \cmark & \xmark \\
    \midrule
    \multicolumn{7}{c}{\emph{Training-free methods}}\\
    OWG~\cite{tziafas2024towards}               & 4-DoF & \cmark & \cmark & \xmark & \xmark & \cmark \\
    ThinkGrasp~\cite{thinkgrasp} & 6-DoF & \cmark & \cmark & \xmark & \xmark & \cmark \\
    GraspMAS~\cite{graspmas}     & 4-DoF & \cmark & \xmark & \xmark & \xmark & \cmark \\
    \red{RoboDexVLM~\cite{liu2025robodexvlm}} & \red{6-DoF} & \red{\cmark} & \red{\cmark} & \red{\cmark} & \red{\xmark} & \red{\xmark} \\
    \midrule
    \rowcolor{gray!12}
    \textbf{\claro{} (ours)}     & 6-DoF & \cmark & \cmark & \cmark & \cmark & \cmark \\
    \bottomrule
  \end{tabular}
  \caption{\textbf{\claro{} combines all six capabilities for general-purpose open-vocabulary grasping in clutter.} Referring grasp
synthesis methods are finetuned on a closed set of objects and operate single-shot (except UnoGrasp).
VLA policies are closed-loop but tied to their training motion distribution and embodiment; existing training-free methods add
open-vocabulary grounding and closed-loop planning but lack long-horizon memory
or non-prehensile manipulation. \cmark/\xmark~denote presence/absence of each
capability.
\emph{Non-Prehensile Manip.}\ is manipulation beyond grasp-and-lift
(e.g.,\ pushing); \emph{Cross-Arm Deploy.}\ denotes methods that deploy across robot arms without re-training.
}
  \label{tab:capabilities}
  \vspace{-10pt}
\end{table*}

Recent works use two frameworks for LDRM. Referring Grasp Synthesis (RGS) methods finetune VLMs to predict grasp poses
directly from an image and a task
instruction, which are then executed by a controller~\cite{tziafas2024towards,Bhat_2026_WACV,thinkgrasp}. These methods are trained on large datasets comprising static RGB-D scenes in which the target is assumed to be visible and graspable from the camera view. This assumption breaks down in dynamic environments when objects are occluded and the robot must intelligently search through multiple manipulation steps. The second framework uses end-to-end Vision Language Action
(VLA) models that learn a mapping from pixels and language to motor actions, and are deployed closed-loop in both simulation and real
hardware~\cite{intelligence2025pi_,openvlaoft,deshpande2026molmob0t}. VLAs enable fine-grained robot control and adaptability to a wide range of embodiments but are often evaluated in-distribution: on objects, scenes, and instructions close to the training data. VLAs report poor performance on unseen and open-vocabulary objects, making them unsuitable for deployment in novel environments~\cite{bhat20253d,yang2026robolab}.
Neither framework suffices: RGS cannot search through clutter, and VLAs do not
generalize beyond their training distribution, yet retrieving a target in
practice requires both. Real environments such as homes, labs, and warehouses
are noisy and cluttered, with targets frequently occluded, so a robust system
must actively declutter the scene through a multi-step, closed-loop process of
planning, grounding, and execution, all while remaining open-vocabulary and
deployable without retraining.

Grounding novel objects and decluttering over multiple steps requires a
system that perceives, plans, and acts in a closed loop. Tool-calling VLM agents, which show strong spatial reasoning beyond conventional
VLMs, are naturally suited to this:
rather than learning a monolithic policy, they invoke skills, observe outcomes, and re-plan~\cite{chen2026spacetools,Song_2026_CVPR}.
A VLM agent in robotics can use a set of skill APIs within a closed loop safety-bounded rulebook, to invoke perception and manipulation functions, observe the environment, and re-plan~\cite{fu2026cap}. Such methods are suitable for clutter manipulation precisely because robot environments are dynamic, the
scene changes after every action, demanding iterative grounding, occlusion reasoning, and re-planning (\cref{fig:ldg-examples}).

We present \claro{} (\textbf{CL}utter-\textbf{A}ware \textbf{R}obot
\textbf{O}rchestrator), a VLM agent that turns a set of perception and
manipulation skills into a general-purpose, open-vocabulary object
grasper. Given a cluttered workspace and a language-specified target, \claro{}
generates planning traces, reasons about occlusions, and executes motion that actively declutters the scene while searching for the target. Once
the object is sufficiently exposed, it executes a feasible grasp to retrieve it. \claro{} requires no environment-specific finetuning and is the only system to combine open-vocabulary grounding, closed-loop planning, long-term memory, $6$-DoF grasping, and non-prehensile decluttering (\cref{tab:capabilities}).

The main contributions of this work are:
\begin{enumerate}
  \item We develop \claro{} (\cref{fig:claro_method}), a VLM agent that plans, declutters, and grasps open-vocabulary objects in tightly cluttered
  workspaces using a robotic arm, and transfers across robot arms and environments without
  retraining.
  \item We benchmark \claro{} against open-vocabulary and trained VLM baselines, across a simulation environment spanning diverse objects, lighting conditions, occlusion levels, and standard image-based RGS benchmarks. \claro{} consistently outperforms all methods, with a relative improvement of $16\%$ in high-clutter success rate over the strongest baseline.
  \item We deploy \claro{} on a robotic arm across 80 trials covering 40 objects in two clutter levels, achieving an average success rate of 88.8\%.
\end{enumerate}

\section{Related Work}

\label{sec:related}

\paragraph{Referring Grasp Synthesis (RGS).}
These methods leverage the joint vision-language embedding space of VLMs to predict a grasp pose directly from an RGB-D image and referring instruction by training on largescale RGB-D object pose and grasping datasets.
CROG used CLIP encoders and a lightweight decoder to train on the OCID-VLG dataset for $4$-DoF grasp pose estimation~\cite{tziafas2023language}.
MapleGrasp introduced mask-guided feature pooling to focus grasp prediction on the referred region while contributing the RefGraspNet $6$-DoF grasping dataset~\cite{Bhat_2026_WACV}. UnoGrasp finetuned a VLM for multi-step obstruction reasoning over cluttered scenes through an occlusion pointing task~\cite{jiao2025obstruction}.  Some methods incorporate chain-of-thought reasoning or attribute and spatial grounding to predict grasps from flexible language instructions~\cite{vcotgrasp,graspcot,yu2025attribute}. While trained RGS models report strong
results on in-domain objects, their performance degrades on out-of-domain objects as models finetuned on a fixed closed-vocabulary datasets lose priors about the open-world objects~\cite{Bhat_2026_WACV}. RGS models are evaluated on static images (where the target is assumed visible and graspable in the scene) rather than in dynamic environments.

\vspace{-15pt}

\paragraph{Vision Language Action (VLA) models.}
VLAs map RGB observations and language to motor actions, trained
on large-scale simulation and tele-operation corpora to provide fine-grained,
closed-loop control~\cite{openvla,intelligence2025pi_}. They excel at contact-rich, articulated and pick-and-place tasks~\cite{deshpande2026molmob0t}, but are typically evaluated
in-distribution (objects, scenes, and task variations close to their
training data) and in low-clutter settings~\cite{bhat20253d}. Current VLAs struggle with limited generalization to
unseen tasks~\cite{yang2026robolab} making them unsuitable for the open-vocabulary grasping problem studied in this paper.

\vspace{-15pt}

\paragraph{VLM agents for open-vocabulary grasping.}
These methods target the limitations of the previous two by pairing a strong open-vocabulary perception module (usually untrained to avoid dataset-specific bias) that grounds the language in 2D bounding boxes or masks with a separate extensively trained grasping module able to handle diverse
geometries. OWG combines a VLM with
segmentation and grasp synthesis, grounding the target via visual prompting and
ranking candidate grasps, while removing obstacles when the target is
blocked~\cite{tziafas2024towards}. ThinkGrasp prompts a VLM to identify the
goal object or the next object likely to reveal it and to suggest grasp locations, which it executes with a segmentation and grasping
network~\cite{thinkgrasp}. GraspMAS uses compositional python program
synthesis, generating executable code that chains vision tools to ground the
referring expression followed by grasping~\cite{graspmas}. \red{RoboDexVLM~\cite{liu2025robodexvlm}
plans a sequence of atomic skills with a VLM, verifies each
execution through depth and fingertip feedback, and re-invokes the planner,
conditioned on earlier episodes, when a skill fails; it
executes with a dexterous hand, and always acts on the target rather than
clearing occluders first. We adapt it to our parallel-jaw setting for comparison
(Supp.~Sec.~3.1).} Other methods add occlusion-aware 3D
detection with active viewpoint planning~\cite{viso} or retrieve past episodes
to guide planning~\cite{lan2025experience}. \claro{} falls in this family: it is a
training-free, closed-loop VLM agent, that incorporates
long-horizon memory and non-prehensile push primitive allowing the robot to
operate in cluttered scenes to find and grasp the target object.

\vspace{-10pt}

\paragraph{\red{VLM agents for general manipulation.}}
\red{Concurrent work extends the agentic framework beyond grasping, such as articulated-object manipulation that requires a broader skill
vocabulary. VoLo~\cite{chen2026volo} treats a VLA model as an interruptible tool,
monitoring its rollout and halting or redirecting it when a subgoal stalls, while
Maestro~\cite{shi2025maestro} composes perception, geometry, and control modules by
writing and revising executable code, repositioning a wrist camera to obtain the
views needed. In both, the target is always visible and reachable. VLMgineer~\cite{gao2025vlmgineer} applies
the same priors offline, placing the model inside an evolutionary search that
co-designs an end-effector and the open-loop trajectory wielding it; its tools are
hardware synthesised for a target action rather than modules invoked over a long
horizon. Articulated manipulation is outside the scope of this work, but the toolshed is the natural place to
extend it: skills such as \texttt{rotate\_handle} or \texttt{grasp\_and\_pull}
would be exposed to CLARO exactly as \texttt{grasp} and \texttt{push} are,
leaving the memory and replanning loop unchanged.}

\section{Methodology}
\label{sec:method}

\begin{figure*}[ht]
  \centering
  \includegraphics[width=\textwidth]{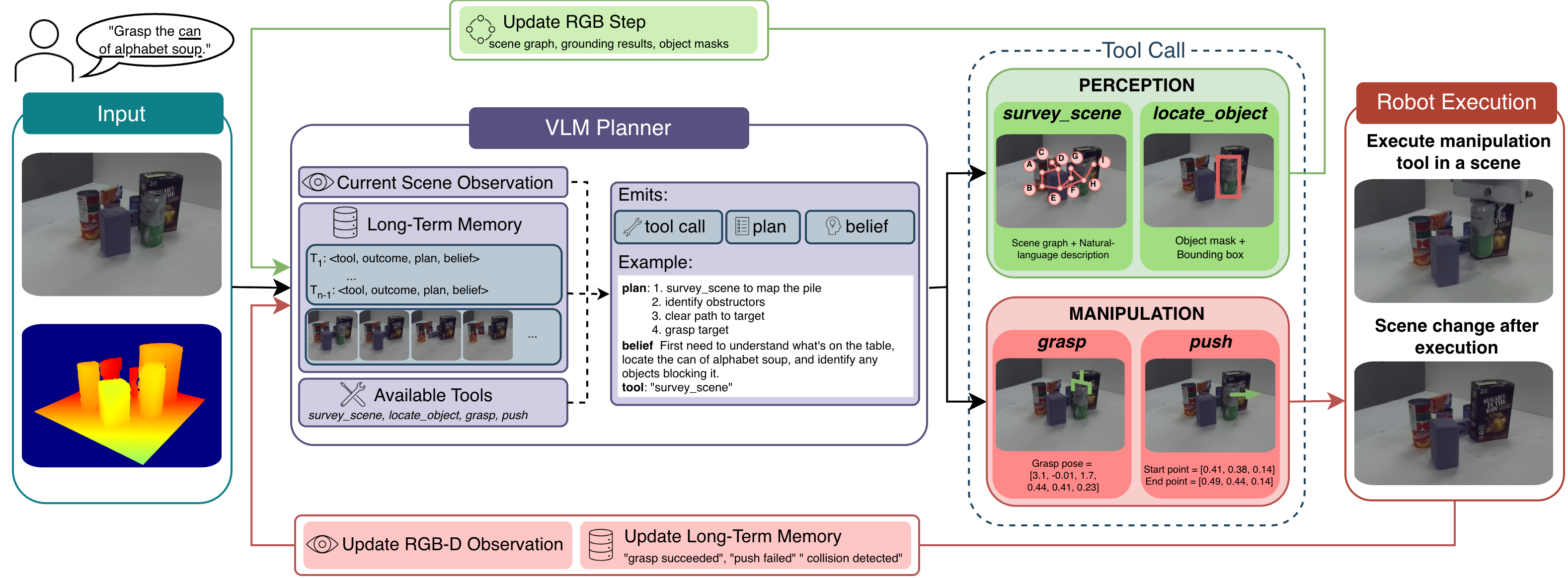}
  \caption{\textbf{\claro{} is a training-free VLM agent that
retrieves a language-specified object from clutter by calling perception and
manipulation tools.} Given an RGB-D observation and an instruction, a VLM
planner conditions on the current scene and a long-term memory
of past images, tool calls, and plan/belief traces to choose the next tool. The
\emph{perception} tools ground the scene while the \emph{manipulation} tools act on it. After each action, the scene changes and the updated
observation is appended to memory.}
\vspace{-10pt}
  \label{fig:claro_method}
\end{figure*}

\claro{} retrieves a referred target by calling perception and manipulation tools in a closed loop
(Fig.~\ref{fig:claro_method}). Open-vocabulary grasping in clutter poses three challenges: (i) An occluded target
requires multi-step search through a scene that changes after every action,
which \claro{} handles by \emph{closed-loop replanning}, re-perceiving the
scene and acting on the observed outcome rather than a fixed plan; (ii) Long-horizon planning introduces drifts, where the agent can ground a rejected object or retry a
failed grasp. \claro{} prevents this through \emph{long-term memory} over
past images, tool calls, and reasoning traces; (iii) Manipulation in clutter involves removing obstructions, yet grasping each can be slow or infeasible. \claro{} adds a \textit{non-prehensile push} tool that can expose a target without grasping.

\subsection{Planner}
\label{sec:method:planner}
\claro{} uses Gemini~3.1~Pro as a VLM planner with access to the current RGB-D
observation and its tools. At each step it selects a tool and emits a short
\emph{plan} and \emph{belief} statement recording its intent and current
understanding of the scene, such as which objects remain occluded or have already
been cleared. Appended to the interaction history, these traces form a part of \claro{}'s
long-term memory and condition every subsequent decision, so the planner avoids the
cyclic re-grounding and repeated failed grasps that arise when steps are treated
independently.

\subsection{Toolshed}
\label{sec:method:tools}
\claro{}'s planner has access to four tools necessary to ground and manipulate objects in each scene.

\vspace{-10pt}

\paragraph{\texttt{survey\_scene}.} Given an RGB image, this tool returns the 2D points of all visible objects and their labels, a top-down scene graph
with directional predicates (\eg\ \emph{left-of},
\emph{behind}), and a description of the scene. It is the
planner's primary visual grounding tool, supplying context needed to identify
the referred target or spot occluders whose removal would reveal hidden objects in the scene. We instantiate this tool with Gemini 3.1 Pro (same as the planner) to maintain a consistent vocabulary through RGB observations.
 \red{By doing so, \texttt{survey\_scene}'s object labels and the planner's own reasoning are drawn from a single concept space (that the VLM learns due to its internet-scale pre-training), which reduces ambiguity in mapping the referred target in text to scene objects (Supp.~Sec.~2.3).}
Planner is encouraged to use this tool after manipulation steps to update visual memory.

\vspace{-10pt}

\paragraph{\texttt{locate\_object}.} This tool takes the RGB image and a text describing the object and returns a 2D bounding box and segmentation mask of the referred object.
\red{Referring text is matched to the corresponding entry in the latest \texttt{survey\_scene}
scene graph, and the 2D point is used to prompt SAM3~\cite{carion2025sam} to
produce a mask and a tightest-fitting bounding box.}
This tool is a precursor to manipulation:
\red{2D mask is lifted to 3D through the depth map which provides a point cloud with one metric point per pixel, $P_{\text{cam}}(v,u)=D(v,u)\,K^{-1}[u,v,1]^{\top}$. The mask is applied to index the object's 3D points, which are then  transformed into the robot base frame by the extrinsics $T_{\text{cam}\to\text{base}}$.}

\vspace{-10pt}

\paragraph{\texttt{grasp}.} Given an RGB-D image and an object mask, this tool
uses GraspGen~\cite{murali2025graspgen} to produce a $6$-DoF grasp for the
object. The planner calls it in two scenarios: when the target is visible and
the robot should retrieve it, or when an occluder can be grasped and removed.

\vspace{-10pt}

\paragraph{\texttt{push}.} This tool produces a non-prehensile motion by using an RGB-D image, object mask, and a direction signal from the planner.
\red{The planner issues \texttt{push} with an object mask, a direction $\in\{$left, right, up, down$\}$ in the \emph{image} frame, and a sweep distance $\in[3,20]$\,cm (default 10); the direction is mapped to the table plane and expressed in the robot base frame through
$T_{\text{cam}\to\text{base}}$. The start point is obtained by ray-tracing from the
object's centroid \emph{opposite} to the push direction, across a 2D occupancy mask of
all detected objects (using 2D points from \texttt{survey\_scene}'s scene graph passed through SAM3~\cite{carion2025sam}), until a gap large enough to lower the closed gripper is
found; if no such gap exists, the tool returns failure, indicating that the clutter
is too dense for a precise push. The end point is the start point displaced by the
sweep distance along the push direction, and the gripper, closed into a flat
paddle, follows the straight segment between them at a fixed height derived from
the object's own point cloud,}
repositioning obstructors that are
better moved aside than grasped due to geometric constraints.

\subsection{Closed-Loop Deployment}
\label{sec:method:loop}
\claro{} runs as a closed loop. Each tool call is executed, the environment updates, and the settled RGB-D observation returns to the planner, which replans based on the outcome. This makes the
search robust to the scene changes, since failed grasps,
shifted obstructors, and newly revealed objects are observed and grounded in the next step rather than ignored.
\red{The resulting tool sequence does not follow a fixed schedule: the planner decides
each call from the current scene and prior outcomes~(Supp.~Fig.~1). We analyse the tool-call sequences \claro{} produces in
Supp.~Sec.~8.}

\vspace{-5pt}

\section{Experiments}
\label{sec:experiments}

We evaluate \claro{} across three complementary settings. First, a
\emph{simulation benchmark} of densely cluttered tabletop scenes measures
end-to-end closed-loop performance under reproducible
conditions. Second, established RGS
benchmarks situate \claro{}'s grounding and grasping against methods evaluated on the standard single-image grasp-detection protocol.
Third, real-robot experiments test generalization across
a distinct embodiment, environment, and camera viewpoint: a broader transfer
setting than conventional evaluations that reuse the same robot-asset
configuration across simulation and real-world experiments. We omit VLA baselines: they generalize poorly out of distribution (79.2\% / 39.2\% for MolmoBot / $\pi_{0.5}$ on unseen objects~\cite{deshpande2026molmob0t}) and are embodiment-locked, requiring retraining that conflicts with our training-free, cross-arm setup. We will release the \claro{} code and simulation benchmark.

\subsection{Simulation Environment and Benchmark}
\label{sec:exp:sim-setup}

\paragraph{Environment.}
Our benchmark is built on the PyBullet physics engine with photorealistic
Blender-based rendering for RGB images as inputs to the perception modules across our baselines. A UR5 arm equipped with a Robotiq
2F-140 parallel-jaw gripper operates over a tabletop observed by a single
static overhead RGB-D camera ($1024\times1024$). The camera is fixed, and every method receives the identical observation at scene initialization. To probe robustness to appearance, each scene is rendered under
two lighting modes: \emph{easy}, a flat ambient source with default shading, and \emph{hard}, randomized indoor HDRI environment maps that
introduce directional shadows, color casts, and varying intensity. Tabletop
textures are randomized per episode. Two simulator-level guards are applied to all methods. (i) A safety check runs an inverse-kinematics feasibility and collision test on every manipulation waypoint; infeasible motion is reported to the method as feedback. (ii) A perception-loop guard terminates an episode if a method issues five consecutive perception calls without an intervening manipulation action, preventing cyclic loops.

\vspace{-10pt}

\paragraph{Asset library.}
Scenes are populated from a collection of $395$ household objects sourced from BOP~\cite{nguyen2025bop} and Digital Twin Catalogue~\cite{Dong_2025_CVPR} model collections, manually filtered to retain
objects whose category and identity are recognizable from an overhead viewpoint without relying on a privileged angle. Each object is annotated with a name, color, and physical properties (collision and mass), and spans categories, including bottles, toys, electronics, food items, kitchenware, figurines, tools, and office supplies.

\vspace{-10pt}

\paragraph{Benchmark.}
The benchmark comprises \textbf{600 episodes}: $3$ random seeds $\times$ $2$
lighting modes $\times$ $100$ episodes each ($3\times2\times100=600$).
Supp.~Fig~5 shows
representative scenes spanning the object, clutter, and lighting diversity.

\vspace{-10pt}

\paragraph{Baselines.}
We compare against \red{six} methods that support language driven grasping: 4 VLM agents from open-vocabulary grasping (OWG~\cite{tziafas2024towards}, ThinkGrasp~\cite{thinkgrasp}, GraspMAS~\cite{graspmas} and \red{RoboDexVLM~\cite{liu2025robodexvlm}}, using GPT 5.4 as the planning VLM), 1 from closed-loop RGS (UnoGrasp~\cite{jiao2025obstruction}), and a naive lower bound implemented using GraspGen~\cite{murali2025graspgen} that executes the highest confidence grasp in the scene, sequentially removing objects without semantic reasoning (denoted by GraspGen*). \red{Baseline implementation details are provided in Supp.~Sec.~3.1.}
\red{We adapt RoboDexVLM by dropping dexterous-hand retargeting and its hand-mounted camera for our fixed overhead view.}
All methods run with a budget of 10 manipulation actions per episode.

\subsection{Evaluation Metrics}
\label{sec:exp:metrics}

\paragraph{Simulation.}
We categorize each episode into two clutter modes, measuring performance under varying complexity. An episode is classified as \emph{high clutter} if the target's visibility is below
$40\%$ and \emph{low clutter} otherwise. Visibility is the ratio of the target's visible pixel area in the
scene to its area in isolation. The 600 benchmark episodes comprise 250 low-clutter and 350 high-clutter scenes. For each clutter mode, we report four metrics; \emph{SR} denotes success rate.

\begin{itemize}
  \item \textbf{Grounding SR} -- the fraction of episodes in which the referred target is localized with a segmentation mask. This measures
  perception accuracy (whether the method can identify the referred
  object) independent of whether it is subsequently grasped or removed.
  \item \textbf{Grasping SR} -- the ratio of successful to attempted
  grasps, averaged over all episodes.
  This metric isolates grasping accuracy, independent of target grounding.
  \item \textbf{Task SR} -- fraction of episodes in which the target is grasped and placed in a receptacle basket. This is the primary metric used to evaluate open-vocabulary grasping.
  \item \textbf{Action Steps} -- average \red{manipulation actions (\texttt{grasp}+ \texttt{push})} per episode; lower value indicates faster solution.
\end{itemize}

\vspace{-15pt}

\paragraph{Static RGS.}
We use the standard grasp detection metric: a predicted grasp is correct if the 2D grasp rectangle overlaps the ground-truth grasp with IOU
$>0.25$ and its orientation error is below $30^\circ$, reported as a percentage~\cite{graspmas}.

\vspace{-5pt}

\begin{table*}[ht]
  \centering
  \footnotesize
  \setlength{\tabcolsep}{5pt}
  \renewcommand{\arraystretch}{1.15}
  \begin{tabular}{l cc cc cc cc}
    \toprule
    & \multicolumn{2}{c}{\textbf{Grounding SR.} (\%) $\uparrow$}
    & \multicolumn{2}{c}{\textbf{Grasping SR.} (\%) $\uparrow$}
    & \multicolumn{2}{c}{\textbf{Task SR.} (\%) $\uparrow$}
    & \multicolumn{2}{c}{\textbf{Action Steps} $\downarrow$} \\
    \cmidrule(lr){2-3}\cmidrule(lr){4-5}\cmidrule(lr){6-7}\cmidrule(lr){8-9}
    \textbf{Method} & Low & High & Low & High & Low & High & Low & High \\
    \midrule
    GraspGen*~\cite{murali2025graspgen}  & 36.2 & 35.4 & 48.8 & \nd{42.2} & 19.2 & 9.4 & 8.8 & 9.2 \\
    GraspMAS~\cite{graspmas}             & 62.8 & 55.0 & 29.2 & 25.8 & 27.2 & 16.7 & \best{1.3} & \best{1.1} \\
    UnoGrasp~\cite{jiao2025obstruction}  & 52.7 & 46.1 & 32.5 & 18.8 & 34.2 & 23.8 & 8.7 & 9.2 \\
    ThinkGrasp~\cite{thinkgrasp}         & 73.8 & 68.8 & 34.5 & 19.2 & 76.2 & 75.3 & 5.4 & 6.4 \\
    OWG~\cite{tziafas2024towards}        & \nd{76.1} & \nd{71.4} & \nd{52.5} & 26.9 & \nd{83.3} & \nd{76.3} & \nd{2.7} & \nd{3.9} \\
    \red{RoboDexVLM~\cite{liu2025robodexvlm}} & \red{70.2} & \red{65.6} & \red{38.0} & \red{28.5} & \red{60.4} & \red{49.2} & \red{6.5} & \red{8.6} \\
    \midrule
    \textbf{\claro{} (ours)} & \best{88.2} & \best{86.3} & \best{60.4} & \best{45.3} & \best{89.5} & \best{88.6} & 3.0 & 4.9 \\
    \bottomrule
  \end{tabular}
  \caption{\textbf{\claro{} achieves the highest task success rate in both low and
high clutter, and unlike the baselines does not degrade as clutter increases.}
GraspMAS reports fewest action steps because its
planner VLM rejects most grasp proposals before execution ($\approx70\%$ of episodes terminate without a single grasp attempt). \best{Bold}:
best; \nd{underline}: second best. GraspGen$^{*}$ is a naive grasp-only baseline that calibrates the task difficulty of our benchmark: simulating a random-pick grasping model only achieves a 19.2\% task SR even in the easier low clutter regime, highlighting the need for occlusion planning and grounding as an important component of open-vocabulary grasping.
}
\vspace{-15pt}
  \label{tab:main}
\end{table*}

\section{Results}
\label{sec:results}

Our evaluation studies three questions. \textbf{(RQ1)} How does
\claro{} perform compared to existing open-vocabulary methods and how does its performance vary with scene complexity and clutter? \textbf{(RQ2)} How do individual components (grounding/grasping) contribute to the overall task performance under varying clutter? and \textbf{(RQ3)} How does \claro{} compare in the standard RGS evaluation criteria on static image benchmarks and annotated grasp poses? We answer RQ1 and RQ2 in simulation (Sec.~\ref{sec:exp:sim}), RQ3 on RGS benchmarks
(Sec.~\ref{sec:exp:rgs}), and isolate the contribution of \claro{}'s components through ablations (Sec.~\ref{sec:exp:abl}).

\subsection{Results on the Simulation Benchmark}
\label{sec:exp:sim}

Table~\ref{tab:main} reports the overall results. \claro{} attains the highest task
success rate in both clutter regimes ($89.5\%$ low, $88.6\%$ high) and, unlike every
baseline, does not degrade significantly as clutter increases (\textbf{RQ1}). The strongest baseline,
OWG, falls from $83.3\%$ to $76.3\%$, a $7$-point drop, while \claro{} yields a relative
improvement of roughly $16\%$ in high clutter compared to baselines. ThinkGrasp reports stable performance across clutter but with lower task success rates ($\sim\!76\%$), and the
finetuned and multi-agent baselines (UnoGrasp, GraspMAS) trail substantially.

\vspace{-10pt}

\paragraph{Grounding.}
In high clutter scenes, \claro{} leads Grounding SR ($86.3\%$ vs.\ OWG's $71.4\%$ and ThinkGrasp's
$68.8\%$), and the gap widens relative to low clutter: every method grounds
less reliably as occlusion increases, but \claro{} degrades the least ($88.2\%
\!\rightarrow\!86.3\%$, a $1.9$-point drop, vs.\ OWG's $4.7$-point drop
$76.1\%\!\rightarrow\!71.4\%$). OWG labels SAM3-segmented objects and
prompts a VLM to select the target, but over-segmentation under clutter misleads this choice. ThinkGrasp prompts LangSAM~\cite{liu2023grounding,carion2025sam}
with VLM-generated text, where naming mismatches often yield no detection. \claro{} improves grounding performance with a shared object vocabulary across planning and grounding. \red{Replacing the pointing tool in \texttt{survey\_scene} with Molmo 2~\cite{Clark_2026_CVPR} (separate object vocabulary between planner and grounder) lowers Grounding SR from $88.2/86.3$ to $77.2/71.6$; the steeper drop in high clutter indicating that preserving object identity across module boundaries matters more as occlusion increases (Supp.~Sec.~2.3).}

\vspace{-5pt}

\paragraph{Grasping.}
\claro{} leads in both clutter regimes and the separation is largest
in high clutter. Grasp reliability also degrades less steeply for
\claro{} ($15$-point drop) than for OWG ($26$-point drop) as clutter increases.
\claro{}'s
$6$-DOF grasping pipeline, combined with the option to push objects into
graspable configurations, lets it sustain reliable grasps where
other baselines collapse.

\vspace{-5pt}

\paragraph{Action efficiency.}
\claro{} uses more steps than OWG ($4.9$ vs.\ $3.9$ in high
clutter), which reflects the additional pushing primitive.
We report this metric to show that \claro{}'s gains come not from manipulating more, but from more focused manipulation through decluttering the target before grasping (Supp.~Fig~4). \red{In successful episodes, \claro{} displaces ${\sim}3.8$ occluders in low clutter and ${\sim}6.6$ in high clutter. These exceed the corresponding Action Steps because a single push
often shifts several neighbouring objects at once. Pushes are excluded from Grasping SR and reported here: \claro{} issues $1.6$ and $2.6$ pushes per episode in low and high clutter.}
CLARO's closed-loop planning averages 34.8 ($\sim$\$0.13) model calls per high clutter episode. We provide cost/latency analysis in the Supp.~Sec~6~.

\vspace{-12pt}

\paragraph{Training-free vs.\ finetuned models.}
UnoGrasp uses a finetuned VLM that performs obstruction reasoning by pointing to
objects to remove. We reproduce this setup by pairing it with SAM3
segmentation and a GraspNet grasp generator (similar to authors' robot experiments). Its Task SR is poor even in low clutter
($34.2\%$), driven by the lowest Grounding SR among all methods: finetuning on
static RGB-D datasets encodes its object and environment biases, which do not
transfer to novel scenes in our benchmark. Training-free methods (OWG, ThinkGrasp, \claro{}) ground with open-world VLM backbones, generalizing better to unseen
objects. The same effect appears on the grasping side: GraspMAS grounds well via
a strong VLM (GPT 5.4), but its grasp model, trained on RGS datasets, yields a low
Grasping SR and hence a low Task SR. Our results show that pairing open-set
detectors with open-set grasp generators is a robust recipe for
open-vocabulary grasping.

\subsection{Results on Static RGS Benchmarks}
\label{sec:exp:rgs}

\begin{table}[t]
  \centering
  \footnotesize
  \setlength{\tabcolsep}{3pt}
  \renewcommand{\arraystretch}{1.15}
  \begin{tabular}{l c c}
    \toprule
    \textbf{Method} & \textbf{OCID-VLG~\cite{tziafas2023language}} & \textbf{GraspAnything++~\cite{vuong2024language}} \\
    \midrule
    OWG~\cite{tziafas2024towards}            & 53.1 & 42.9 \\
    ViperGPT~\cite{surismenon2023vipergpt} & 44.0 & 57.0 \\
    ThinkGrasp~\cite{thinkgrasp} & 48.5 & 38.4 \\
    GraspMAS~\cite{graspmas} & 62.0 & 68.0 \\
    \textbf{\claro{} (ours)} & \textbf{67.2} & \textbf{76.5} \\
    \bottomrule
  \end{tabular}
  \caption{\textbf{On standard benchmarks, \claro{} exceeds baselines despite being a training-free agent.} Success rate ($\%$) measured under the RGS protocol (\cref{sec:exp:metrics}). \claro{} attains the best score on both
OCID-VLG and GraspAnything++ benchmarks. OWG, ThinkGrasp, GraspMAS are evaluated with our implementation; ViperGPT is quoted from the original paper.}

  \label{tab:rgs_benchmarks}
\end{table}

The simulation benchmark highlights \claro{}'s interactive abilities, which are not measured in
static single-image benchmarks. To verify that our VLM agentic design does not compromise basic grounding
competence, we evaluate on two RGS benchmarks
(\cref{tab:rgs_benchmarks}) under the standard grasp-detection metric. \claro{} attains the highest success rate on both OCID-VLG
($67.2\%$) and GraspAnything++ ($76.5\%$).

\subsection{Ablations}
\label{sec:exp:abl}

\begin{table}[t]
  \centering
  \footnotesize
  \setlength{\tabcolsep}{5pt}
  \renewcommand{\arraystretch}{1.2}
  \begin{tabular}{l cc cc}
    \toprule
    & \multicolumn{2}{c}{\textbf{Task SR} (\%) $\uparrow$}
    & \multicolumn{2}{c}{\textbf{Action Steps} $\downarrow$} \\
    \cmidrule(lr){2-3}\cmidrule(lr){4-5}
    \textbf{Variant} & Low & High & Low & High \\
    \midrule
    \textbf{\claro{} (full)}          & \best{89.5} & \best{88.6} & 3.0 & 4.9 \\
    \;\,w/o \texttt{push} tool       & 77.4 & 60.0 & 6.6 & 7.5 \\
    \;\,w/o long-term memory          & 71.2 & 52.9 & \textbf{1.9} & \textbf{3.8} \\
    \;\,w/o closed-loop replanning    & 74.9 & 56.6 & 3.1 & 4.1 \\
    \red{\;\,w/o VLM based planning} & \red{62.5} & \red{54.8} & \red{5.4} & \red{6.2} \\ %
    \midrule
    \;\,GraspNet as grasper& 82.1 & 80.9 & 4.5 & 5.0 \\
    \;\,GPT 5.4 as planner & 83.8 & 78.2 & 3.3 & 5.2 \\
    \red{\;\,Qwen3.5-397B-A17B as planner} & \red{86.0} & \red{85.2} & \red{4.3} & \red{7.0} \\
    \midrule
    \bottomrule
  \end{tabular}
  \caption{\textbf{\claro{}'s core capabilities drive its clutter
robustness: removing long-term memory, closed-loop replanning, or
pushing drops high-clutter Task SR upto $36$ pp.}
Replacing the grasp generator (GraspGen~$\rightarrow$~GraspNet) or the VLM planner (Gemini 3.1 Pro~$\rightarrow$~GPT~5.4 \red{or the open-weight Qwen3.5-397B-A17B}) causes modest drops.
\red{\emph{w/o VLM based planning} runs with a hard-coded loop where the VLM only decides when to grasp/push.}
}
\vspace{-17pt}
  \label{tab:ablation}
\end{table}

We ablate the capabilities that distinguish \claro{} from existing open-vocabulary grasping methods: non-prehensile pushing, long-term memory, closed-loop replanning, and
\red{VLM based planning, which we replace with a hard-coded
\texttt{survey}$\rightarrow$\texttt{locate}$\rightarrow$\texttt{grasp/push}
sequence} (VLM decides grasp or push). We also ablate over \claro{}'s system design choices, GraspGen for GraspNet as the
\texttt{grasp} tool and Gemini~3.1~Pro \red{or the open-weight
Qwen3.5-397B-A17B} for GPT~5.4  as the planner/grounder in \texttt{survey\_scene}, over the same $600$ episodes
(\cref{tab:ablation}). Without \texttt{push}, high-clutter Task
SR falls $28.6$ points ($88.6\!\rightarrow\!60.0$) and Action Steps rise as the
agent falls back on repeated grasps to clear obstructors. Without long-term
memory, the planner oscillates,
frequently calling perception tools rather than manipulating the scene until our
perception-loop guard terminates the episode ($1.9/3.8$ Action Steps); removing
closed-loop replanning reduces performance ($56.6$), as the agent retries
grasps whose failure it no longer observes. \red{A hard-coded fixed planner, instead of a VLM, is the largest
drop of all ($62.5/54.8$) showing that \claro{}'s tool orchestration rather than the manipulation primitives strengthens the
performance.} Varying system design choices affects performance marginally, with every
variant still exceeding all baselines (\cref{tab:main}): GraspNet gives $82.1/80.9$, still outperforming OWG
and ThinkGrasp, while GPT~5.4 \red{and the open-weight Qwen3.5-397B-A17B based planner} reports $83.8/78.2$ and \red{$86.0/85.2$} success rates indicating that \claro{}'s performance is not tied to a particular tool choice.

\vspace{-5pt}

\section{Real-Robot Experiments}
\label{sec:realrobot}

\begin{figure}[ht]
  \centering
  \setlength{\tabcolsep}{2pt}
  \renewcommand{\arraystretch}{0.6}
  \includegraphics[width=\linewidth]
  {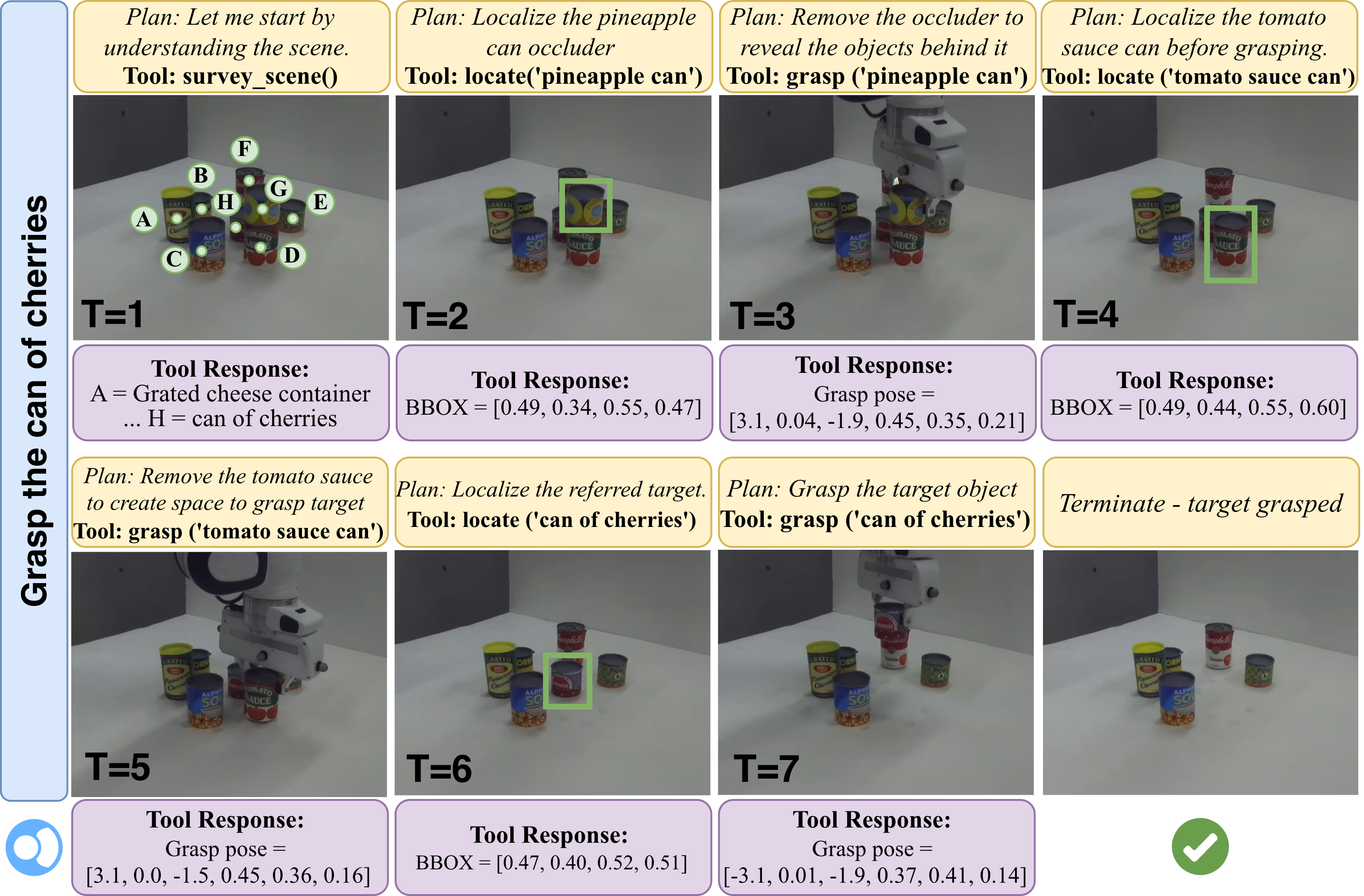}
  \caption{\textbf{A robot rollout of \claro{} retrieving an occluded target.} The task is to grasp
the can of cherries, initially hidden behind other objects.
\claro{} first calls
\texttt{survey\_scene} to map the pile (T$=$1), then alternates between
\texttt{locate\_object} and \texttt{grasp} to remove the occluders that block the target (T$=$2--5), before
localizing and grasping the can of cherries once exposed (T$=$6--7).
}

  \label{fig:real-robot}
\end{figure}

\paragraph{Setup.}
We used a Franka arm, mounted on a tabletop and equipped with a parallel gripper. A fixed ZED2 RGB-D camera captures images of the tabletop in front-view. We use 40 objects to create cluttered scenes for our experiments (\cref{fig:real-robot}). We compare \claro{} with OWG and ThinkGrasp.

\vspace{-10pt}

\paragraph{Protocol.}
Each method is evaluated on 16 tasks, with 5 independent trials. Out of the 16 tasks, 8 correspond to low clutter scenes ($\le$7 objects well spaced) and 8 correspond to high clutter ($\geq$8 objects and densely placed). Thus, we run 80 experiments per method and report the overall task success rate, average number of action steps with range of deviation. In each task, the robot is given a natural-language target object and an initial RGB-D observation of the cluttered tabletop. Each method operates in closed loop until the target is successfully retrieved, the method declares failure, or its grasp budget is exhausted.  All methods are allowed up to five grasp attempts per episode.

\vspace{-5pt}

\begin{table}[ht]
  \centering
  \footnotesize
  \setlength{\tabcolsep}{2.5pt}
  \renewcommand{\arraystretch}{1.2}
  \begin{tabular}{l cc cc}
      \toprule
      & \multicolumn{2}{c}{\textbf{Task SR} (\%) $\uparrow$}
      & \multicolumn{2}{c}{\textbf{Action Steps} $\downarrow$} \\
      \cmidrule(lr){2-3}\cmidrule(lr){4-5}
      \textbf{Method} & Low & High & Low & High \\
      \midrule
      OWG~\cite{tziafas2024towards}
        & 67.5 $\pm$ 35.4 & 22.5 $\pm$ 32.8
        & 3.4 $\pm$ 1.2 & \nd{4.6 $\pm$ 0.9} \\
      ThinkGrasp~\cite{thinkgrasp}
        & \nd{70.0 $\pm$ 33.8} & \nd{50.0 $\pm$ 35.5}
        & \best{3.2 $\pm$ 1.3} & \best{4.3 $\pm$ 1.1} \\
      \midrule
      \textbf{\claro{} (ours)}
        & \best{97.5 $\pm$ 7.1} & \best{80.0 $\pm$ 23.9}
        & \nd{3.3 $\pm$ 1.1} & 5.4 $\pm$ 1.8 \\
      \bottomrule
    \end{tabular}
  \caption{\textbf{Results on a Franka arm (n = 40 per cell: 8 low/high clutter scenes × 5 trials)}. Task SR is reported with 95\% confidence intervals from a scene-clustered bootstrap. CLARO outperforms both baselines in high clutter (paired Wilcoxon signed-rank over 8 scenes: p = 0.008 vs OWG, p = 0.031 vs ThinkGrasp).}
  \label{tab:realrobot}
  \vspace{-10pt}
\end{table}

\paragraph{Results.}
Table~\ref{tab:realrobot} shows that \claro{} achieves the highest task success
at both clutter levels, improving over the strongest baseline by $27.5$ points in
low clutter ($97.5$ vs.\ $70.0$) and $30.0$ points in high clutter ($80.0$ vs.
$50.0$). \cref{fig:real-robot} shows a sample task from our experiments. The low-to-high clutter gap is wider on the real robot than in simulation, for
two reasons. First, the camera observes the table from an oblique front view: this aids language grounding but makes manipulation harder,
as feasible grasps require accurate $3$D alignment whereas top-down grasps
tolerate slight misalignment. Second, our manually designed scenes induce
stronger occlusion than the simulator, including cases where the target is not
visible from the initial view.

The baseline ranking reverses relative to simulation:
ThinkGrasp outperforms OWG in real-world experiments. This is a consequence of the
oblique viewpoint. OWG's grasp generator predicts $4$-DoF grasps, which need to be orthographically reconstructed to $6$-DoF grasps introducing additional noise.
ThinkGrasp's $6$-DoF grasps remain feasible from the
same view. All methods deteriorate in high clutter, where heavy occlusion and long-horizon planning make the task difficult. Stronger planners and richer manipulation tools are promising directions to
close this gap.

\vspace{-5pt}

\section{Conclusion}
\label{sec:conclusion}

\vspace{-5pt}

We presented \claro{}, a training-free, closed-loop VLM agent for
open-vocabulary grasping in clutter. By equipping a VLM planner with a compact
set of perception and manipulation tools while preserving the full interaction history as long-horizon memory,
\claro{} searches, declutters, and retrieves a language-specified target without
any environment or embodiment specific finetuning. On our 600-episode simulation benchmark, \claro{} achieves the highest task success in both
low and high clutter and, unlike all baselines, does not degrade as clutter
increases, improving over the strongest competing method by $\sim16\%$ in
high clutter. Real-robot experiments on a Franka arm establish strong sim2real generalizability, demonstrating \claro{} can transfer to different robot arms, camera, and object sets.

\red{\paragraph{Practical takeaways for VLM agents in robotics.}
Long-horizon memory is effective when provided to VLMs as structured outcome state rather than visual replay. Removing memory costs
$35.7$\,pp of high-clutter Task SR, and decomposing the retained context into its
textual and visual channels (Supp.~Sec.~4) locates that benefit: discarding every
past frame while keeping the tool-result trace costs $8.2$\,pp, whereas keeping
the frames and discarding the text costs $26.9$.
Second, a cheap non-prehensile primitive is worth more than
a stronger grasp model, and its value grows with occlusion: removing \texttt{push}
costs $12.1$\,pp in low clutter but $28.6$ in high. Since a grasp-only toolset appears adequate in
light clutter, VLM-agent evaluations should be stratified by target visibility
rather than reported in aggregate. Third, reliable grounding does not imply
reliable manipulation: from low to high clutter \claro{}'s Grounding SR falls only
$1.9$\,pp ($88.2\!\to\!86.3$) while per-attempt Grasping SR falls $15.1$
($60.4\!\to\!45.3$), and on the real robot grasp and push execution account for
$80.4\%$ of failure events against $19.6\%$ for segmentation (Supp.~Tab.~4).}

\vspace{-10pt}

\paragraph{Limitations and future work.}
\claro{} operates from a single camera, adopted for fair comparison
with our single-view baselines. This limits point-cloud density for grasp
estimation and prevents the agent from choosing where to observe. Active view selection could further improve grounding and grasping
under occlusion.
\claro{}'s planner may select any tool within its safety and budget constraints, and it can perform redundant actions, such as running \texttt{survey\_scene} after a grasp that leaves the scene unchanged. \red{Re-grounding only the objects near the contact point can reduce such redundancies.}
We leave metrics that reduce this planning overhead to
future work.
\red{Retained frame history also dominates prompt-token growth; since the structured outcome trace carries most of the memory benefit (Supp.~Sec.~4), pruning frames under a bounded window or summarization policy is a natural extension to longer horizons.}
\claro{} manipulates objects without object specific physical
constraints, such as avoiding fragile items. Extending to object-aware
manipulation preferences is a promising direction for safety-critical settings.

{
    \small
    \bibliographystyle{ieeenat_fullname}
    \bibliography{main}
}

\end{document}


\maketitle

{\renewcommand{\thefootnote}{}\footnotetext{$^{*}$Equal contribution.}}


\section{Overview of Supplementary Material}
\label{sec:supp_overview}

This supplementary provides the implementation detail, additional experiments,
and analyses that support the main paper's claims.
\cref{sec:supp_prompts} gives
the full planner and \texttt{survey\_scene} prompts, a precise specification of
\claro{}'s long-term memory, and an annotated episode trace.
\cref{sec:baseline_fidelity} documents how each baseline was reproduced in our
unified environment.
\cref{sec:memory_decomposition} decomposes the memory ablation to identify which
component drives the $36$\,pp (main paper, table 4) effect. \cref{sec:supp_failures} presents a failure-mode taxonomy for the
real-robot runs, \cref{sec:supp_cost} reports the per-episode cost and latency
referenced in the main text, and \cref{sec:supp_benchmark} details the benchmark
instructions, object pool, occlusion threshold, and seeding.
\red{
\cref{sec:supp_toolflow} analyzes \emph{how} \claro{} uses its tools: when the planner pushes rather than grasps, and how \claro{}'s decisions compare with human
judgment.}
A supplementary
video (\cref{sec:supp_video}) shows real-robot successes and failures.

\section{Planner and \texttt{survey\_scene} Prompts with an Annotated Episode Trace}
\label{sec:supp_prompts}

\subsection{Planner System Prompt}

CLARO uses a VLM planner to choose the next high-level action from the current RGB observation and the accumulated interaction history. The planner is given access to four tools: \texttt{survey\_scene}, \texttt{locate\_object}, \texttt{grasp}, and \texttt{push}. We provide the full system prompt used in our simulation experiments in \cref{fig:planner_system_prompt}.

\begin{figure*}[t]
  \centering
  \begin{tcolorbox}[
    width=0.98\textwidth,
    title={Planner System Prompt},
  ]
\begin{Verbatim}
You are an embodied robotic grasping agent operating a UR5 arm with a Robotiq 2F-140 gripper over a cluttered tabletop. Your goal: grasp a specified target object and place it in the basket/tray.

The conversation that follows is your interaction history with the simulator: you call a tool, receive its result, and decide the next action.

== ENVIRONMENT ==
- UR5 + Robotiq gripper, top-down RGB camera, basket/tray for placed objects.
- The robot, gripper, stand, and tray are not visible in the rendered view -- you see only the table and objects.
- Each user turn includes an attached image of the CURRENT scene. Earlier images in this conversation are still visible to you and represent the state at earlier points in time -- use them to compare "before" vs "after" any manipulation.
- Images are NEVER annotated; all perception results and object IDs are given as text inside the user turn.
- The scene is extremely cluttered. The target object is likely partially or fully occluded by other objects. You must strategically remove or push obstructing objects to reach and grasp the target.

== TOOLS (call exactly ONE per turn) ==

1. survey_scene()
   Generates a detailed scene graph: every visible object with its name, rich description (color, shape, branding), 2D position, and spatial relations (on_top_of, partially_occluded_by, next_to). Also produces a scene_summary.
   - Use this at the START of an episode to build holistic scene understanding.
   - Use this AFTER every manipulation to update your understanding of the scene changes.

2. locate_object(object_name)
   A specialized grounding model that precisely locates ALL visible instances of a named object. Returns referred object's SAM-segmented masks suitable for manipulation.
   - Use this when you need to MANIPULATE a specific object (grasp or push).
   - If the object is not found, status=not_found is returned -- the object is likely occluded.

3. grasp(object_id)
   Robot grasps the object with marker `object_id` (from the latest locate_object) and drops it in the basket/tray.
   - REQUIRES a locate_object result whose marker IDs are still valid.
   - Marker IDs become INVALID after any grasp or push -- always run locate_object again before acting.
   - This is both the manipulation tool AND the way to complete the task. Grasp the target object to succeed.

4. push(object_id, direction, distance_cm)
   Slide an object sideways with a closed-gripper sweep.
   - `direction` must be one of: "left", "right", "up", "down" (relative to the image).
   - `distance_cm` (optional) is the sweep distance in centimetres, between 3 and 20. Default is 10.
   - Use push to clear obstructors that are on top of or next to the target.
   - Pushing has a higher success rate than grasping. Use it to declutter efficiently.

== BUDGET ==
- Perception (survey_scene, locate_object) is FREE -- no cost.
- Only grasp and push calls consume the budget. You have a maximum of 10 attempts per episode.
- Plan wisely: you need sufficient grasp attempts remaining once the target is finally exposed. 

== OUTPUT FORMAT -- STRICT JSON ONLY ==
Respond with EXACTLY ONE JSON object. No prose, no markdown, no code fences.

Required fields:
- "tool": one of "survey_scene", "locate_object", "grasp", "push"
- "belief": 1-2 sentences on why THIS action now
- "plan": stepwise plan to reach the target (required turn 1; optional after)

Tool-specific fields:
- locate_object: "object_name"
- grasp: "object_id"
- push: "object_id", "direction", optionally "distance_cm" (3-20, default 10)

Examples:
{"tool": "survey_scene", "plan": "1. survey_scene to map the pile, 2. identify obstructors, 3. clear path to target, 4. grasp target", "belief": "First turn -- need to understand what's on the table and where the target might be."}
{"tool": "locate_object", "object_name": "large blue box on top", "belief": "Scene graph shows blue box is on_top_of the target area -- need marker ID to remove it."}
{"tool": "grasp", "object_id": 2, "belief": "Object 2 is the obstructor blocking the target. Removing it should expose the target underneath."}
{"tool": "push", "object_id": 3, "direction": "left", "distance_cm": 12, "belief": "Object 3 is next to the target and partially covering it. Pushing left should clear the path."}
{"tool": "grasp", "object_id": 1, "belief": "Object 1 is the target (orange baby spoon). Grasping it completes the task."}
\end{Verbatim}
  \end{tcolorbox}
  \caption{
  Full planner system prompt used by \claro{} in simulation experiments.
  }
  \label{fig:planner_system_prompt}
\end{figure*}

\subsection{\texttt{survey\_scene} Prompt}
\label{sec:supp_survey_scene_prompt}

The \texttt{survey\_scene} tool uses a VLM to convert the current top-down RGB observation into a structured scene graph. This scene graph gives the planner a shared object vocabulary for later grounding calls: visible objects are assigned names, descriptions, normalized image locations, and spatial relations such as \texttt{on\_top\_of}, \texttt{partially\_occluded\_by}, and \texttt{next\_to}.
The full \texttt{survey\_scene} system prompt is shown in \cref{fig:survey_scene_prompt}.

\begin{figure}[!t]
  \centering
  \small
  \begin{tcolorbox}[
    title={\texttt{survey\_scene} System Prompt},
  ]
\begin{lstlisting}[
    basicstyle=\ttfamily\fontsize{6pt}{6pt}\selectfont,
    showstringspaces=false,
    numbers=none,
    frame=none
  ]
You are an expert visual scene analyst. Given a top-down image of
objects on a table, produce a detailed scene graph in YAML format.
For EACH separate physical object, provide:
  - obj_id: sequential integer starting at 1
  - name: most specific name you can read from visible labels; if no label, use a short descriptive name
  - desc: 1-2 sentence rich description (color, shape, material, any text/branding visible, approximate size relative to other objects)
  - point_norm: [row, col] object centre on a normalised [0.0, 1.0] grid (row 0=top, col 0=left). 
  - position_description: short phrase (e.g. 'top-left', 'center', 'far right')
  - relations: list of spatial relations to OTHER objects, each as [relation_type, other_obj_id]. Use ONLY these relation types:
      on_top_of             -- this object is resting on top of other_obj_id
      under             -- this object is resting under other_obj_id
      partially_occluded_by -- part of this object is hidden by other_obj_id
      next_to               -- this object is adjacent to other_obj_id
      left-of,              -- this object is resting on the left of other_obj_id
      right-of,              -- this object is resting on the right of other_obj_id
      behind                -- this object is resting on behind other_obj_id
    Only include relations you are confident about. An empty list is fine.
If you see multiple identical objects, list each one separately.
Do NOT include the table surface, floor, shadows, or background.
Also provide a top-level 'scene_summary' field: a 2-3 sentence free-text description of the overall scene layout, notable occlusions, and anything that might be hidden.
Return STRICT YAML. No markdown code fences.
Example:
scene_summary: "7 objects visible. Cluttered pile in center..."
objects:
  - obj_id: 1
    name: "soup can"
    desc: "small red can with 'ALPHABET' text"
    point_norm: [0.32, 0.49]
    position_description: "center"
    relations:
      - [next_to, 2]
  - obj_id: 2
    name: "yellow bottle"
    desc: "cylindrical yellow squeeze bottle with white cap"
    point_norm: [0.16, 0.13]
    position_description: "top-left"
    relations:
      - [partially_occluded_by, 3]
\end{lstlisting}
  \end{tcolorbox}
  \caption{
  Full \texttt{survey\_scene} system prompt used by \claro{} to generate a
  structured scene graph from the current RGB observation with the \texttt{survey\_scene} tool.
  }
  \label{fig:survey_scene_prompt}
\end{figure}

This prompt is designed to make perception and planning share the same object-level representation. The planner reasons over the object names, descriptions, and spatial relations produced by \texttt{survey\_scene}, and then passes the same natural-language object references to \texttt{locate\_object} for precise mask grounding. This shared vocabulary reduces mismatches between high-level planning and low-level grounding, especially in cluttered scenes where the target may be partially occluded and must be identified through its relations to surrounding objects.

\subsection{\red{Effect of the Shared Object Vocabulary}}
\label{sec:supp_vocab}
\red{To isolate the contribution of the shared object vocabulary, we decouple
naming from grounding while holding the remainder of the pipeline fixed: the
Gemini~3.1~Pro component supplying object identity to grounding is replaced by
the open-weight model Molmo 2-8B, with SAM3,
the planner, and all 600 benchmark episodes unchanged. Grounding SR falls from
88.2/86.3 to 77.2/71.6 (low/high clutter). Failure inspection shows cross-model
naming drift: Molmo can reduce a specific reference such as ``coffee can with
MASTER written on it'' to the coarser ``blue can,'' making nearby objects harder
to disambiguate. The drop is larger in high clutter, where less of the target
is visible: with its distinguishing features occluded, a coarse label such as
``blue can'' is more readily matched to a co-present object than to the target
itself. }

\subsection{Long-Term Memory}
\label{sec:supp_long_term_memory}

At each step, \claro{} maintains two buffers: the VLM-visible conversation
memory \texttt{conversation} and the structured execution log
\texttt{action\_history}. The memory field passed back to the VLM is
\texttt{conversation}. Before the VLM call at step \(t\),
\texttt{conversation} contains the previous turns
\(\{1,\ldots,t-1\}\), including prior RGB observations
\(\texttt{rgb\_1},\ldots,\texttt{rgb\_\{t-1\}}\), user prompts, and
VLM JSON responses. The current call additionally receives the current RGB
observation \(\texttt{rgb\_t}\) and prompt \(\texttt{user\_prompt\_t}\).

After the VLM is called at step \(t\), \claro{} appends the current user turn
and the VLM response to \texttt{conversation}:
\begin{verbatim}
{"role": "user",
 "text": user_prompt_t,
 "images": [rgb_t]}

{"role": "assistant",
 "text": json(vlm_response_t),
 "images": []}
\end{verbatim}
Therefore, after step \(t\), the VLM-visible memory contains the sequence of
past observations, prompts, and responses up to that step. Let
\(u_i=\texttt{user\_prompt\_i}\) and
\(v_i=\texttt{vlm\_response\_i}\). Then
\[
\begin{aligned}
\mathcal{C}_t
&=
\{(\texttt{rgb\_1},u_1,v_1),\ldots,
(\texttt{rgb\_t},u_t,v_t)\},\\
&\text{where } \mathcal{C}_t
\text{ denotes } \texttt{conversation} \text{ after step } t.
\end{aligned}
\]

The VLM response is a JSON tool-call record. It stores the selected tool, the
belief string, the plan string when emitted, and the tool-specific arguments:
\begin{verbatim}
vlm_response_t =
{
  "tool": tool_t,
  "belief": belief_t,
  "plan": plan_t,
  "...tool_specific_args...": "..."
}
\end{verbatim}
For example, \texttt{locate\_object} includes \texttt{object\_name},
\texttt{grasp} includes \texttt{object\_id}, and \texttt{push} includes
\texttt{object\_id}, \texttt{direction}, and optionally
\texttt{distance\_cm}. Thus, the tool call, tool arguments, belief trace, plan
trace, current image, and prompt are stored in \texttt{conversation} and
remain visible in later VLM calls.

After the selected tool is executed, \claro{} appends one tool-specific record
to \texttt{action\_history}. This record stores the executed tool, belief string, success/failure outcome, and tool-specific execution details:
\begin{verbatim}
{"step_number": t,
 "tool": tool_t,
 "belief": belief_t,
 "success": success_t,
 "...execution_details...": "..."}
\end{verbatim}

At step \(t+1\), \claro{} reads the latest entry
\texttt{action\_history[-1]} and serializes selected fields into a
\texttt{TOOL RESULT} block. This block contains the tool name, the
succeeded/failed outcome, and selected tool fields such as object name, push
direction, and push distance. For perception tools, the prompt may also include
the current scene graph or the latest \texttt{locate\_object} result when
available.

The \texttt{TOOL RESULT} block is inserted into
\texttt{user\_prompt\_\{t+1\}}, and that prompt is then appended to
\texttt{conversation}. Consequently, older tool outcomes persist in
long-term memory because previous prompts containing \texttt{TOOL RESULT}
blocks remain in \texttt{conversation}, together with the corresponding
past images and VLM responses.

\subsection{Worked Example of Long-Term Memory-Guided Recovery}

\begin{figure}[ht]
  \centering
  \setlength{\tabcolsep}{2pt}
  \renewcommand{\arraystretch}{0.6}
  \includegraphics[width=\linewidth]
  {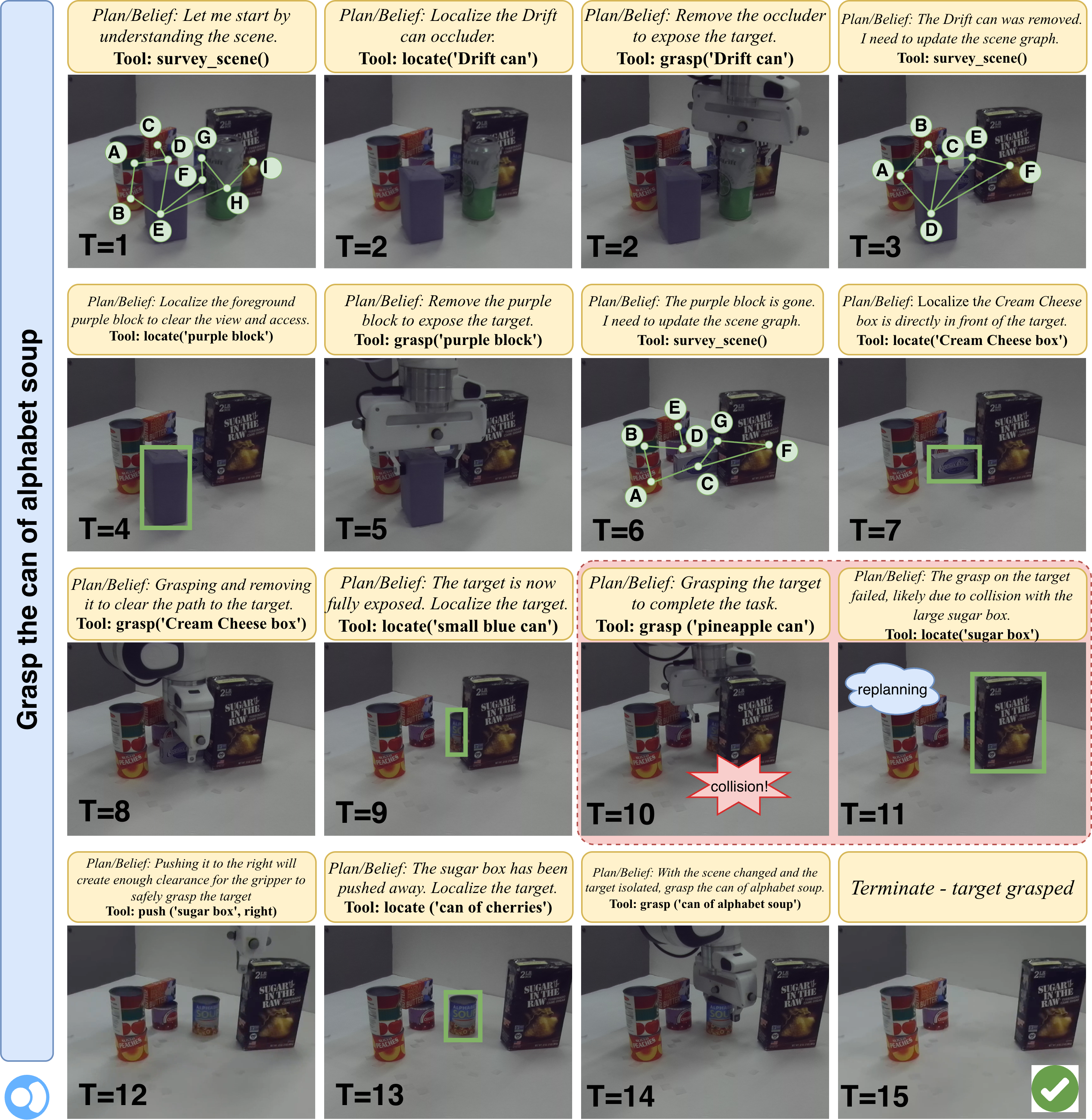}
  \caption{
  Worked high-clutter success episode for grasping the can of alphabet soup. The dashed red region highlights the
  memory-guided recovery: after the target grasp fails, \claro{} does not repeat the same failed grasp. Instead, it pushes the sugar box away, re-localizes and grasps the target successfully.
  }
  \label{fig:supp_success_episode_transcript}
\end{figure}

Fig.~\ref{fig:supp_success_episode_transcript} shows a worked high-clutter
success episode for the instruction ``grasp the can of alphabet soup.'' The
episode illustrates how \claro{} uses the accumulated interaction history to
recover from a failed target grasp. After the first target grasp fails, the next
turn does not repeat the same grasp. Instead, the planner uses the previous
failure outcome to identify the nearby sugar box as the remaining constraint,
pushes it away, re-localizes the target after the scene change, and then grasps
the target successfully.

\section{Baseline Reproduction Fidelity}
\label{sec:baseline_fidelity}

To ensure a fair comparison, we integrate all baselines into a unified simulation environment with identical evaluation protocols. Table~\ref{tab:baseline_fidelity} summarizes the pipeline components and integration approach for each method.

\begin{table*}[t]
\centering
\footnotesize
\caption{Baseline reproduction fidelity. All methods are evaluated in the same PyBullet environment (UR5 + Robotiq 2F-140) with identical seeds, episode sequences, manipulation budgets (10 actions), and success criteria.}
\label{tab:baseline_fidelity}
\begin{tabular}{lccc}
\toprule
\textbf{Method} & \textbf{VLM} & \textbf{Grounding} & \textbf{Grasping} \\
\midrule
CLARO (Ours) & Gemini 3.1 Pro & SAM3  & GraspGen 6-DOF \\
OWG & GPT-5.4 & SAM3 & GR-ConvNet 4-DOF  \\
ThinkGrasp & GPT-5.4 & LangSAM & GraspNet 6-DOF  \\
GraspMAS & GPT-5.4 & VLPart + SAM3 & RAGT 4-DOF  \\
UnoGrasp & UnoGrasp (fine-tuned) & SAM3 & GraspNet 6-DOF \\
\red{RoboDexVLM} & \red{GPT-5.4} & \red{LangSAM} & \red{AnyGrasp 6-DOF} \\
GraspGen* & None (random) & SAM3 & GraspGen 6-DOF \\
\bottomrule
\end{tabular}%
\end{table*}

\subsection{Integration Details}

\paragraph{OWG.}
The original OWG pipeline uses an open-vocabulary detector for object grounding followed by GR-ConvNet for 4-DOF planar grasp generation. In our integration, we replace the open-vocabulary detector with SAM3 since it is a more recent and better segmentation model (box-prompted from VLM-predicted regions) while preserving the same GR-ConvNet grasp generator and VLM planning logic. The 4-DOF grasp formulation is maintained as in the original work. We use the same SAM3 model in CLARO too.

\paragraph{ThinkGrasp.}
ThinkGrasp employs VLM-based scene reasoning followed by LangSAM grounding and GraspNet 6-DOF grasp generation. Our integration preserves this full pipeline, adapting only the execution layer for the PyBullet UR5 workspace. LangSAM and GraspNet operate with their released weights.

\paragraph{GraspMAS.}
GraspMAS uses a multi-agent VLM architecture (observer, planner, coder) with VLPart for part-level detection and RAGT for 4-DOF grasp generation. We reproduce the full pipeline using the released VLPart and RAGT weights, adapting only the scene interface for our environment. For consistency, we use SAM3 to obtain segmentation masks.

\paragraph{UnoGrasp.}
UnoGrasp uses a fine-tuned Qwen2.5-VL model that predicts 2D coordinates indicating which object to grasp. The original pipeline pairs this with GroundedSAM (text-prompted using the predicted object name) and GraspNet on the full scene point cloud, filtering grasps by the segmentation mask. In our integration, we use the released checkpoint and pair the VLM output with SAM3 (point-prompted using the predicted coordinates) and GraspNet on the masked point cloud. This re-pairing is more direct: it uses the model's coordinate output without an intermediate text-to-detection step and we verified that the checkpoint reproduces the published benchmark numbers by running the authors' evaluation code on the UnoBench test set separately.

\paragraph{\red{RoboDexVLM.}}
\red{RoboDexVLM sequences atomic skills with a VLM planner over LangSAM grounding and executes them with a five-finger dexterous hand. Its grasps originate as parallel-gripper proposals from AnyGrasp, which a kinematic
retargeting stage converts into hand configurations. Our adaptation removes that
stage rather than replacing it: the Robotiq 2F-140 consumes the proposals in their original form. We further replace its hand-mounted RGB-D camera with the fixed overhead RGB-D camera shared by all other methods, equalizing the observation across baselines. Skill sequencing, memory retrieval, grounding, and replanning are preserved.}

\paragraph{Reproduction verification.} Where published numbers exist on the benchmarks we use, our reproductions match them: OWG and GraspMAS fall within 1 pp of their reported OCID-VLG and GraspAnything++ scores (main paper, Tab. 3), and the UnoGrasp checkpoint reproduces its published UnoBench results under the authors' own evaluation code. ThinkGrasp and RoboDexVLM report on neither benchmark, so their numbers here are our measurements and are not externally anchored.

\subsection{Fair Evaluation Protocol}

All baselines are evaluated under identical conditions:
\begin{itemize}
    \item Same PyBullet environment: UR5 robot, Robotiq 2F-140 gripper, fixed table geometry
    \item Same object assets: unified object set with realistic meshes and physics
    \item Same seed: deterministic episode sequences across all methods
    \item Same budget: 10 manipulation actions per episode
    \item Same camera: fixed overhead view
    \item Same success criterion: target object detected inside the receptacle basket
\end{itemize}

\noindent This protocol ensures that performance differences reflect algorithmic capability rather than environmental advantages.

\begin{table}[t]
  \centering
  \footnotesize
  \setlength{\tabcolsep}{5pt}
  \renewcommand{\arraystretch}{1.2}
  \begin{tabular}{l cc cc}
    \toprule
    & \multicolumn{2}{c}{\red{\textbf{GPT 5.4}}}
    & \multicolumn{2}{c}{\red{\textbf{Gemini 3.1 Pro}}} \\
    \cmidrule(lr){2-3}\cmidrule(lr){4-5}
    \red{\textbf{Task SR} (\%)} & \red{Low} & \red{High} & \red{Low} & \red{High} \\
    \midrule
    \red{OWG}        & \red{83.3} & \red{76.3} & \red{55.2} & \red{49.8} \\
    \red{ThinkGrasp} & \red{76.2} & \red{75.3} & \red{80.8} & \red{78.2} \\
    \bottomrule
  \end{tabular}
  \caption{\red{\textbf{Swapping the planner VLM in baselines.} Replacing
  GPT~5.4 with Gemini~3.1~Pro breaks OWG's set-of-mark prompting and yields
  only a marginal gain for ThinkGrasp; neither approaches \claro{}
  (89.5/88.6).}}
  \label{tab:supp_planner_swap}
\end{table}
\subsection{\red{Planner VLM Swap in Baselines}}
\label{sec:supp_planner_swap}
\red{
To separate the contribution of \claro{}'s planner model from that of its
architecture, we replaced the GPT~5.4 planner with Gemini~3.1~Pro in
the two strongest baselines and re-evaluated on the full benchmark
(Table~\ref{tab:supp_planner_swap}). The two baselines react in opposite
directions. OWG's Task SR \emph{falls} from 83.3/76.3 to 55.2/49.8 (low/high
clutter): OWG uses set-of-mark visual prompting, in which the planner sees the
image overlaid with colored masks and must select the target by index, and
Gemini~3.1~Pro associates the referred object with its overlay less reliably
than GPT~5.4, so misselections propagate to grounding and grasping.
ThinkGrasp, which prompts the planner with the raw RGB image (as \claro{}
does), improves marginally from 76.2/75.3 to 80.8/78.2. Both variants remain
well below \claro{} (89.5/88.6), so the choice of planner VLM alone does not
explain \claro{}'s advantage.
}

\section{Memory-Content Decomposition}
\label{sec:memory_decomposition}

A key result from our ablation study is that removing long-term memory drops high-clutter Task SR by 36\,pp (88.6\%$\to$52.9\%). This raises a natural follow-up: \emph{which component} of the accumulated context drives this effect? We decompose the agent's memory into three orthogonal channels and evaluate each in isolation.

\subsection{Memory Channels}

At each reasoning step, CLARO's conversation history contains interleaved turns:

\begin{enumerate}
    \item \textbf{User turns}: an RGB observation image \emph{and} a structured textual block comprising tool results (scene graphs, grounding outcomes, grasp success/failure), task reminders, and budget/warning signals.
    \item \textbf{Assistant turns}: the VLM's own prior tool-call decisions (JSON-structured action selections with reasoning traces).
\end{enumerate}

\noindent We isolate three subsets:

\begin{itemize}
    \item \textbf{Text-Only}: All textual history (tool results, scene graphs, budget state) but no prior images. The VLM reasons from its accumulated belief state without visual grounding of earlier steps.
    \item \textbf{Images-Only}: All prior RGB observations but text replaced with a minimal placeholder. The VLM sees how the scene changed but receives no explicit outcomes.
    \item \textbf{Actions-Only}: Only the VLM's own prior tool-call JSON responses. It recalls what it decided but lacks both observations and outcomes.
\end{itemize}

\subsection{Results}

\begin{table}[t]
  \centering
  \footnotesize
  \setlength{\tabcolsep}{5pt}
  \renewcommand{\arraystretch}{1.2}
  \begin{tabular}{l cc cc}
    \toprule
    & \multicolumn{2}{c}{\textbf{Task SR} (\%) $\uparrow$}
    & \multicolumn{2}{c}{\textbf{Action Steps} $\downarrow$} \\
    \cmidrule(lr){2-3}\cmidrule(lr){4-5}
    \textbf{CLARO Memory Variant} & Low & High & Low & High \\
    \midrule
    Full (default)       & \textbf{89.5} & \textbf{88.6} & 3.0 & 4.9 \\
    Text-Only            & 86.8 & 80.4 & 3.1 & 5.2 \\
    Images-Only          & 76.3 & 61.7 & 2.4 & 4.0 \\
    Actions-Only         & 73.1 & 55.8 & 2.1 & 3.9 \\
    No-Memory            & 71.2 & 52.9 & 1.9 & 3.8 \\
    \bottomrule
  \end{tabular}
  \caption{\textbf{Memory decomposition.} The textual belief trace (tool results, scene graphs) accounts for the majority of the memory benefit. Images-Only and Actions-Only perform close to No-Memory, confirming that structured outcome signals, not raw visual replay, drive multi-step planning. Low action steps in degraded variants reflect stalling (repeated failures) rather than efficiency.}
  \label{tab:memory_decomp}
\end{table}

Table~\ref{tab:memory_decomp} reveals clear stratification:

\paragraph{Text-Only retains most of the benefit.}
Removing images from history costs only 2.7\,pp in low clutter and 8.2\,pp in high clutter, retaining $>$90\% of the memory advantage. The structured belief trace: scene graphs, grasp outcomes, budget warnings provides sufficient context for coherent multi-step planning. The agent correctly avoids re-attempting failed grasps, tracks which occluders have been removed, and adapts strategy under budget pressure, all without re-observing prior frames.

\paragraph{Images-Only recovers little over No-Memory.}
Retaining only prior RGB frames yields marginal improvement over having no memory at all (+5.1\,pp low, +8.8\,pp high). Visual memory helps the VLM notice scene changes (objects disappearing, new surfaces exposed) but without explicit tool-result text, it cannot reliably distinguish successful grasps from failed attempts that left the scene unchanged. The reduced action steps (2.4/4.0 vs.\ 3.0/4.9 for Full) indicate that the agent stalls, cycling through perception calls without productive manipulation, rather than acting efficiently.

\paragraph{Actions-Only is near-equivalent to No-Memory.}
Knowing what was \emph{attempted} without knowing outcomes or current visual state provides almost no planning benefit (+1.9\,pp low, +2.9\,pp high over No-Memory). The only observable effect is mild repetition-avoidance, the agent occasionally picks a different object than its last attempt.

\subsection{Discussion}

This decomposition reveals that CLARO's primary memory advantage is through the structured belief trace (scene graphs, tool success/failure signals, budget state) that enables efficient multi-step planning under occlusion. Visual memory provides a complementary change-detection signal but is not the dominant driver.

This finding connects to two other results. First, it explains why the \textsc{w/o closed-loop replanning} ablation (main paper, Table 4) also causes a large drop: replanning depends on tool-result \emph{text} to know what changed, so removing feedback is functionally similar to removing textual memory. Second, it has practical implications for deployment under context-window constraints: preserving the textual plan trace should be prioritized over retaining all prior frames, and investing in high-quality tool-result formatting yields outsized returns.

\section{Failure-Mode Taxonomy}
\label{sec:supp_failures}

\subsection{Real World Experiments}

We analyze failed \claro{} episodes by counting failure-inducing steps during the episode rather than assigning a single root cause to each episode.

\begin{table}[t]
  \centering
  \footnotesize
  \setlength{\tabcolsep}{5pt}
  \renewcommand{\arraystretch}{1.2}
  \begin{tabular}{lcc}
    \toprule
    \textbf{Failure source} & \textbf{Count} & \textbf{Fraction} \\
    \midrule
    Perception / SAM-mask quality & 10 & 19.61\% \\
    Grasp generation or execution & 22 & 43.14\% \\
    Push planning or execution & 19 & 37.25\% \\
    \midrule
    \textbf{Total failure events} & \textbf{51} & \textbf{100.00\%} \\
    \bottomrule
  \end{tabular}
  \caption{\textbf{Real-robot failures are dominated by manipulation-level
  budget consumption rather than a single semantic planning error.} Grasp generation/execution
  accounts for the largest share of failure events, followed by push-related
  failures and SAM-mask quality issues.}
  \label{tab:supp_failure_taxonomy_real}
\end{table}

\begin{itemize}
    \item \noindent \textbf{Perception / SAM-mask quality:} This category captures cases where the selected point was associated with an incomplete or
noisy SAM mask. These mask errors can affect subsequent manipulation:
for grasping, a partial mask can produce a grasp on a part of the object and
lead to collision; for pushing, mask noise in the oblique-view in robot trials can shift the perspective of an object causing the push motion to start from a bad location.
    \item \noindent \textbf{Grasp generation or execution:} This category covers cases where the intended object was localized, but the
grasp step consumed manipulation budget without success. Examples include low-confidence or
unavailable grasp generation, collision during execution, or a grasp that fails
to lift the object. We did not observe inverse kinematics based infeasibility as a dominant failure in
the real-robot runs, since the workspace was constrained to a reachable region.
    \item \noindent \textbf{Push planning or execution:}
This category covers push actions that did not de-clutter the scene or made the
scene setup more complex for subsequent manipulation. Examples include pushes that increased occlusion, failed pushes during execution, or consumed action
budget without creating useful clearance for the target grasp.
\end{itemize}

Overall, the largest fraction of failure-inducing events comes from grasp
generation or execution, followed by push-related failures and SAM-mask quality.
This suggests that the highest impacting factor to improving real-robot performance is to reduce compounding grasping failures. For videos showing failure in robot experiments, please refer to the supplementary material.

\subsection{\red{Planner-Level Failure Analysis}}
\label{sec:supp_planner_failures}
\red{
The taxonomy above concentrates on perception- and manipulation-level events;
here we ask how often the \emph{planner's reasoning itself} causes failure. We manually reviewed every failed \claro{} episode in one 100-episode pass of the
benchmark (one seed, easy lighting), excluded failures attributable to grounding or grasp execution, and classified the remainder into three modes:
\begin{itemize}
    \item \textbf{Semantic grounding hallucination} (5 episodes): the planner
    never resolves the target during the episode.
    \item \textbf{High-level reasoning error} (3 episodes): the target is
    resolved but budget is committed to an ineffective strategy, such as pushing
    non-blocking distractors while a visible occluder remains.
    \item \textbf{Planning loop} (2 episodes): the planner repeats
    non-productive actions on the correct target without changing approach.
\end{itemize}

These account for 10 episodes, distributed as 5, 3, and 2 respectively. No single reasoning
pathology dominates, indicating that planner errors are diffuse rather than traceable to one systematic weakness. Isolating them at scale would require a planner-level metric and a large-scale human evaluation of planning
trajectories, which we leave for future work.}

\begin{figure*}[t]
  \centering
  \setlength{\tabcolsep}{2pt}
  \renewcommand{\arraystretch}{0.7}
  \includegraphics[width=\linewidth]
  {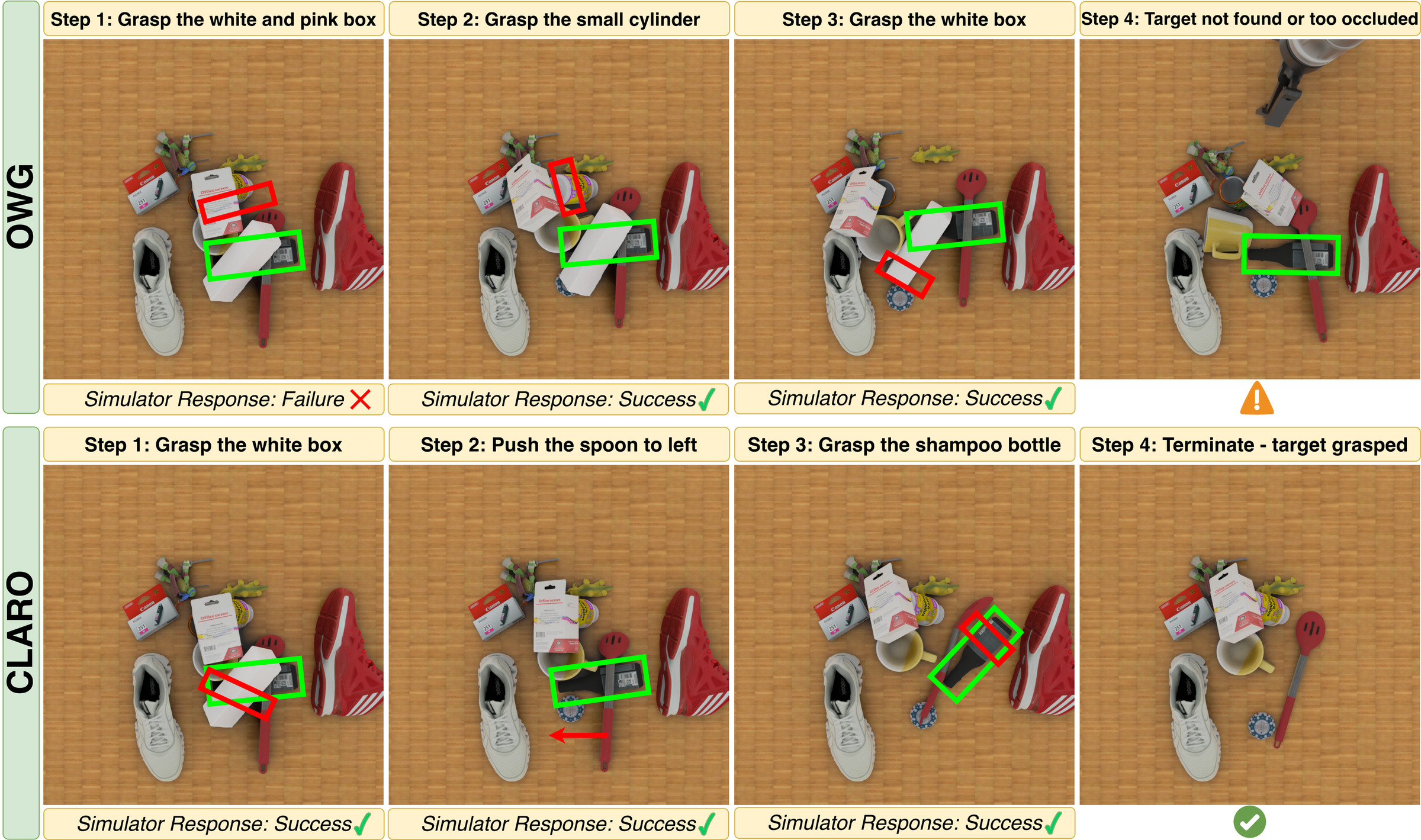}
  \vspace{-5pt}
  \caption{\textbf{\claro{} plans focused actions to grasp the target, while the best baseline (OWG) removes the wrong objects and fails to recover.} The task is to ``grasp the black shampoo bottle", which is marked by the green bounding box. Each panel shows the top-view RGB observation immediately before a manipulation, with the chosen primitive overlaid in red and the simulator's per-step outcome below.
  }
  \label{fig:traces}
  \vspace{-5pt}
\end{figure*}

\section{Cost and Latency Breakdown}
\label{sec:supp_cost}

\subsection{\red{Model Calls and Cost}}
\label{sec:supp_cost_calls}

CLARO's closed-loop planning uses multiple VLM calls per episode: one
\emph{planner} call per reasoning step (deciding which tool to invoke),
plus one additional VLM call for each \texttt{survey\_scene} invocation
(the scene-graph perception prompt).
Local models (SAM3, GraspGen) incur no API cost.  Table~\ref{tab:cost}
summarises the per-episode breakdown measured over 100 high-clutter episodes (easy lighting).

\begin{table}[h]
\centering
\caption{Per-episode model-call and cost breakdown (based on Gemini 3.1 Pro
Preview, June 2026 pricing).}
\label{tab:cost}
\small
\begin{tabular}{lrrr}
\toprule
\textbf{Call Type} & \textbf{Mean} & \textbf{Median} & \textbf{Cost Share} \\
\midrule
Planner       & 21.0 & 17 & 74\% \\
\texttt{survey\_scene} &  13.8 &  6 & 26\% \\
\midrule
\textbf{Total model calls}      & \textbf{34.8} & \textbf{23} & --- \\
\midrule
\textbf{API cost / episode}     & \textbf{\$0.13} & \$0.09 & --- \\
\bottomrule
\end{tabular}
\end{table}

\paragraph{\red{Baseline inference budgets.}}
\red{For a like-for-like comparison we measured per-episode VLM usage for the
two strongest closed-loop baselines under the same Gemini~3.1~Pro backend
\claro{} uses (Table~\ref{tab:supp_budget}).
CLARO's retained
RGB history dominates token consumption, and the \textsc{Text-Only} variant
(\cref{sec:memory_decomposition}) discards it entirely, reducing cost to
$\sim\!\$0.052$ per episode while attaining 86.8/80.4 low/high clutter Task SR, above both OWG
(83.3/76.3) and ThinkGrasp (76.2/75.3) evaluated at their full budgets. CLARO \textsc{Text-Only} variant reports slightly higher VLM calls per episode than the full version, but since it uses no historical RGB frames its overall token utilization and thus cost is lower. \textsc{Text-Only} variant still retains a $\approx$ 2.5x inference budget over the baselines.}

\begin{table}[t]
  \centering
  \footnotesize
  \setlength{\tabcolsep}{6pt}
  \renewcommand{\arraystretch}{1.2}
  \begin{tabular}{l rr}
    \toprule
    \red{\textbf{Method}} & \red{\textbf{VLM calls / ep.}} & \red{\textbf{API cost / ep.}} \\
    \midrule
    \red{OWG}        & \red{6.2}  & \red{\$0.023} \\
    \red{ThinkGrasp} & \red{5.6}  & \red{\$0.021} \\
    \midrule
    \red{\claro{} (full)} & \red{34.8} & \red{\$0.13} \\
    \red{\claro{} (\textsc{Text-Only} memory)} & \red{35.2} & \red{\$0.052} \\
    \bottomrule
  \end{tabular}
  \caption{\red{\textbf{Per-episode inference budget, all methods under the
Gemini~3.1~Pro planner.} All methods are priced at the same Gemini~3.1~Pro
Preview rates (June 2026), so cost differences reflect call count and tokens per
call rather than model pricing; \claro{}'s retained RGB history makes its calls
more image-heavy than the baselines'. Discarding those frames
(\textsc{Text-Only}) cuts \claro{}'s cost by more than half at an unchanged call
count.}}
  \label{tab:supp_budget}
\end{table}
\subsection{Latency Breakdown}
\label{sec:supp_latency}

VLM inference accounts for only 22.5\% of total wall-clock time; the
remaining 77.5\% is simulation infrastructure overhead: primarily
Blender's EEVEE keyframe rendering (used solely for realistic high quality images as input to baselines) and PyBullet
physics stepping with conservative settle-time checks.
\red{The average wall-clock time per simulation episode in \claro{}, 142.5s, is therefore a property of the simulation harness
(dominated by rendering and physics settling that
every method pays for its own calls), not an execution latency of the
architecture: the VLM round-trip averages $\approx$32\,s per episode, and the
rendering/settling component is absent in real-robot deployment.}

\paragraph{Accuracy--latency trade-off.}
CLARO's sequential, closed-loop decision-making (median 17 planner
calls, mean 21) is a deliberate design choice: each step conditions on
the \emph{updated} scene state after the previous action, enabling
reactive re-planning when a grasp fails, an object shifts unexpectedly,
or a push reveals previously-occluded geometry.  This stands in contrast
to open-loop planners that commit to a full action sequence upfront.

\red{The additional calls are not themselves the source of the performance advantage: despite costing less than half as much per episode, the \textsc{Text-Only} variant still outperforms every baseline (\cref{sec:supp_cost_calls}).}

Reducing the per-episode call count (e.g.\ via batched multi-step
planning, plan caching across similar scenes, or distillation of the
VLM policy into a lightweight reactive controller) is a promising
avenue for future work.

\section{Benchmark and Object-Category details}
\label{sec:supp_benchmark}

\begin{figure*}[ht]
  \centering
  \setlength{\tabcolsep}{2pt}
  \renewcommand{\arraystretch}{0.6}
  \includegraphics[width=\linewidth]
  {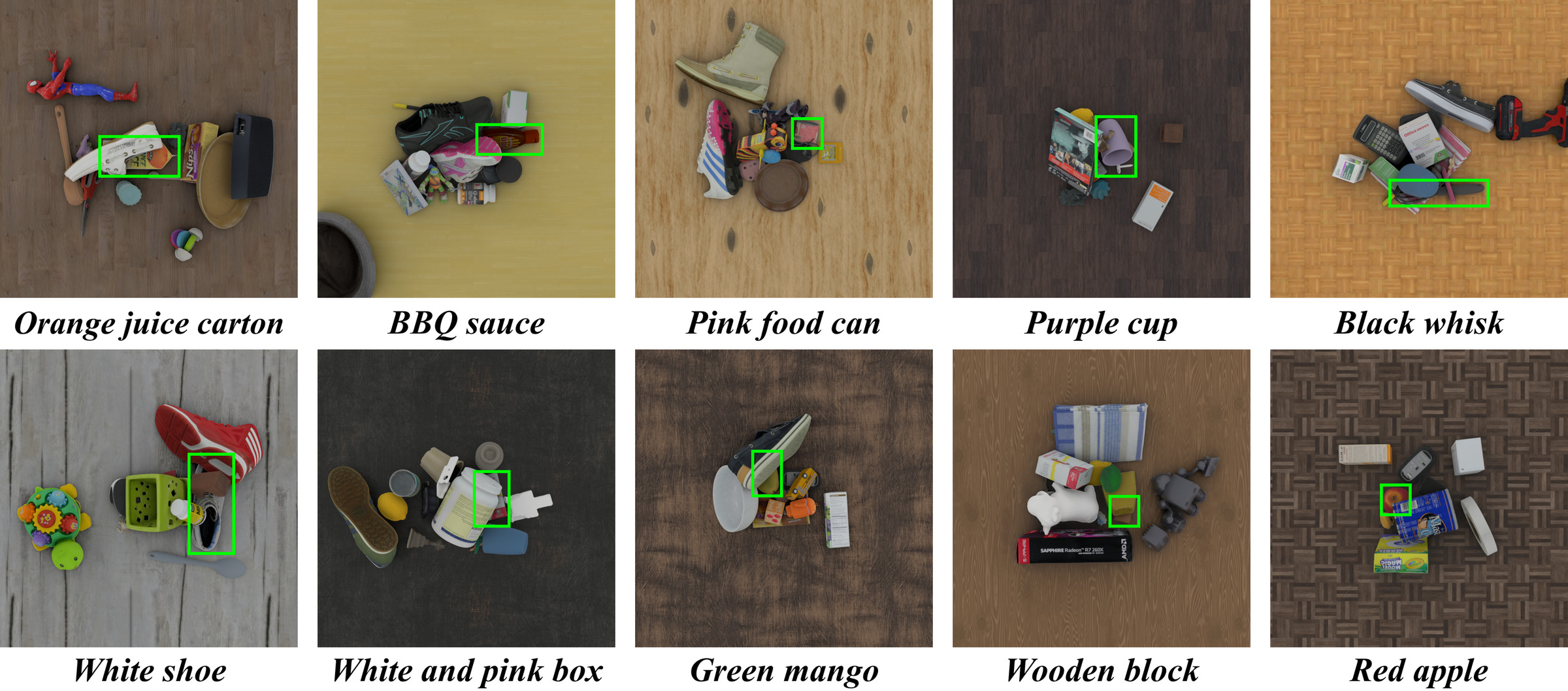}
  \caption{\textbf{Our simulation benchmark spans diverse object categories, materials, table
textures, and clutter levels, with targets ranging from prominent to heavily
occluded.} Each panel shows a scene with its referred target marked in
a green bounding box. }
  \label{fig:benchmark}
  \vspace{-15pt}
\end{figure*}

\subsection{Task Instruction Format}
\label{sec:supp_instructions}

Task instructions in the benchmark follow a fixed template:

\begin{center}
\texttt{Grasp the "\{object\_name\}"}
\end{center}

\noindent
where \texttt{object\_name} is the human-verified common name assigned
to the target object (e.g., ``pink computer mouse'', ``canned peaches'',
``small ceramic vase'').  This templated format was chosen deliberately
to isolate the \emph{physical} challenge of cluttered grasping from
the \emph{linguistic} challenge of diverse instruction parsing:

\begin{itemize}
\item \textbf{Names are appearance-grounded}: Each object name references
  visually discriminative attributes (color, size, material, or function)
  that distinguish it from co-present distractors.  Examples:
  ``pink dumbbell'' (vs.\ green dumbbell), ``navy boat shoe'' (vs.\
  white boat shoe), ``rectangular supplement box'' (vs.\ supplement
  bottle).

\end{itemize}

\paragraph{\red{Prompt-template generalization study.}}
\red{To test sensitivity to instruction phrasing beyond the fixed template, we
replaced ``Grasp the \{object\}'' with instructions sampled from a 50-prompt
set: 25 \emph{direct} queries that name the target with varied phrasing
(``Pick up the \{object\}'', ``I need the \{object\}, fetch it for me'') and 25
\emph{indirect} queries that specify it only by color, shape, or spatial
relation (``Retrieve the small red object on the left'', ``Grab the
cylindrical item next to the blue box''). Each generated instruction was
manually verified to identify a unique target in its scene. \claro{} along with the two strongest baselines are
re-evaluated over the same 600 benchmark episodes
(Table~\ref{tab:supp_diverse_prompts}, averaged over both clutter regimes).

On direct queries every method stays within half a point of its
original-template score, so phrasing variation alone is not a bottleneck.
Indirect queries separate the methods. Grounding SR falls by 11.8\,pp for OWG,
7.2\,pp for ThinkGrasp, and 4.0\,pp for \claro{}, and the losses propagate
downstream: Task SR falls by 9.2\,pp for OWG and 9.0\,pp for ThinkGrasp,
against 3.6\,pp for \claro{}. \claro{}'s indirect-query Task SR (85.4) remains
above both baselines' \emph{direct}-query scores.

The difference reflects how each method resolves the referring expression.
OWG's planner selects targets from an image overlaid with colored masks, which
alters apparent object color and obscures the neighbouring context that
referring expressions depend on. ThinkGrasp's grounding backend (LangSAM)
struggles with complex referring queries. \claro{} instead uses
\texttt{survey\_scene} to recover scene-level attributes and spatial
predicates, then passes the selected object reference to
\texttt{locate\_object} for precise grounding.}

\begin{table}[t]
  \centering
  \footnotesize
  \setlength{\tabcolsep}{4pt}
  \renewcommand{\arraystretch}{1.2}
  \begin{tabular}{l cc cc}
    \toprule
    & \multicolumn{2}{c}{\red{\textbf{Grounding SR} (\%)}}
    & \multicolumn{2}{c}{\red{\textbf{Task SR} (\%)}} \\
    \cmidrule(lr){2-3}\cmidrule(lr){4-5}
    \red{\textbf{Method}} & \red{Direct} & \red{Indirect} & \red{Direct} & \red{Indirect} \\
    \midrule
    \red{ThinkGrasp} & \red{71.2} & \red{64.0} & \red{75.6} & \red{66.6} \\
    \red{OWG}        & \red{73.6} & \red{61.8} & \red{79.4} & \red{70.2} \\
    \red{\claro{}}   & \red{\textbf{87.2}} & \red{\textbf{83.2}} & \red{\textbf{89.0}} & \red{\textbf{85.4}} \\
    \bottomrule
  \end{tabular}
  \caption{\red{\textbf{Prompt-template generalization} over the 600 benchmark
  episodes with 25 direct and 25 indirect instruction templates, averaged over
  both clutter regimes. \claro{}'s indirect-query Task SR exceeds both
  baselines' direct-query scores.}}
  \label{tab:supp_diverse_prompts}
\end{table}

\subsection{Object Annotation Pipeline}
\label{sec:supp_annotations}

The benchmark object pool comprises \textbf{395 objects} drawn from two
mesh datasets:

\begin{enumerate}
\item \textbf{BOP Objects} (220 meshes): house-hold and industrial items.
\item \textbf{DTC Objects} (776 meshes from DTC scans): smaller items
  including toy food, figurines, fruits, dumbbells, calculators.
\end{enumerate}

\noindent
From the combined pool of 996 meshes, we selected 395 based on
graspability (fits within the Robotiq 2F-140 gripper's 14\,cm aperture)
and visual distinctiveness.

\paragraph{VLM-assisted labelling.}
Object metadata was generated using \textbf{Gemini 3.1 Flash} with an
8-view rendered turntable of each mesh (for objects with rendered
thumbnails) or UV texture maps (for texture-only assets).  The VLM was
prompted to produce: (1) a concise common name, (2) dominant color(s),
(3) shape descriptor, (4) graspability assessment, and (5) confidence
score.  Each annotation was then \textbf{manually verified} by two
annotators who accepted, corrected, or rejected names.  The final fields
per object are:

\begin{center}
\small
\texttt{\{name, color, category, graspable, size\_cm\}}
\end{center}

\noindent
Targets were selected to span
diverse sizes (3--20\,cm), textures (glossy, matte, transparent),
and geometries (cylindrical bottles, flat mice, irregular figurines).

\subsection{Object-Category Distribution}
\label{sec:supp_categories}

\begin{figure*}[h]
\centering
\includegraphics[width=\linewidth]{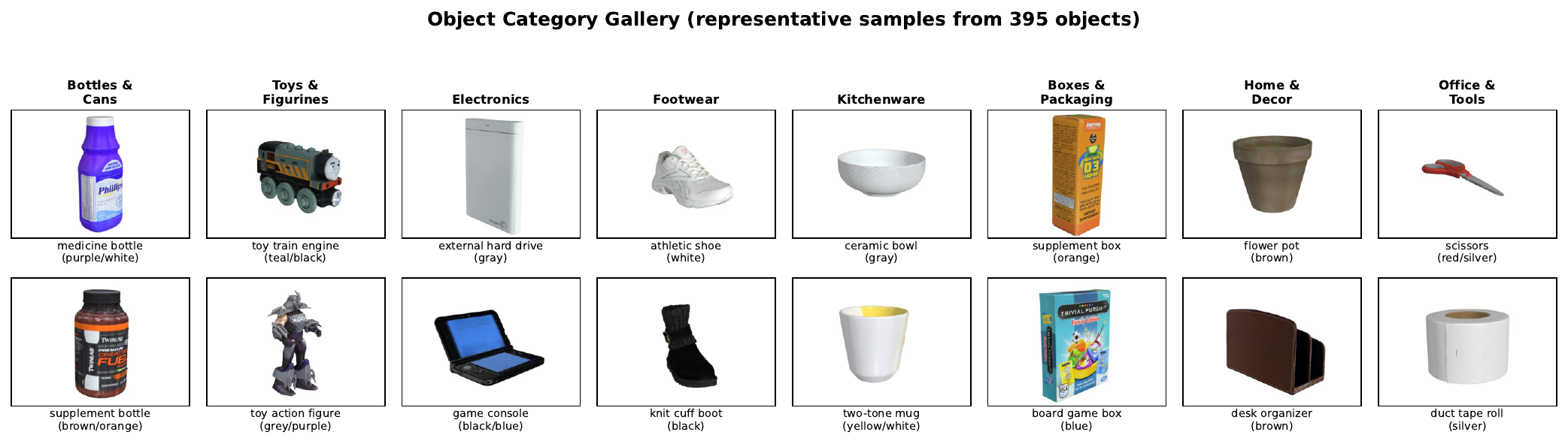}
\caption{Representative objects from each category.  The pool spans
diverse geometries, materials (glossy, matte, textured), sizes, and visual complexity.}
\label{fig:gallery}
\end{figure*}

Table~\ref{tab:obj_categories} shows the object-category breakdown.
Figure~\ref{fig:gallery} presents representative rendered thumbnails
from each category.

\begin{table}[h]
\centering
\caption{Object pool by super-category.}
\label{tab:obj_categories}
\small
\begin{tabular}{lr}
\toprule
\textbf{Category} & \textbf{Total} \\
\midrule
Toys \& Figurines      & 86 \\
Bottles \& Cans        & 37 \\
Boxes \& Packaging     & 56 \\
Electronics            & 28 \\
Footwear               & 49 \\
Kitchenware            & 27 \\
Home \& Decor          & 22 \\
Office \& Tools        & 16 \\
Fruits                 & 11 \\
Sports                 &  4 \\
Other                  &  59 \\
\midrule
\textbf{Total}         & \textbf{395} \\
\bottomrule
\end{tabular}
\end{table}

\subsection{Occlusion Threshold Justification}
\label{sec:supp_threshold}

The benchmark reports results split by a \textbf{40\% visibility
threshold} on low-high clutter scenes: episodes where the target's visibility is less than 40\% are categorized as high-clutter, and low-clutter if otherwise.

\paragraph{Empirical calibration.}
This threshold was set by running a \emph{naive GraspGen-only baseline}
(no planning, no pushing: simply attempt a 6-DOF grasp on the target's
visible mask) across 200 scenes with varying occlusion levels.  At 60\% (40\% visibility)
occlusion, the baseline's direct-grasp success rate drops below 5\%,
confirming that objects above this threshold are effectively unreachable
without deliberate decluttering.

\subsection{Episode Generation and Seed System}
\label{sec:supp_seeds}

Each experiment uses a single base seed that
deterministically controls:
\begin{itemize}
\item Object selection: 15 objects sampled from the 395-object pool
  (one designated target, 14 distractors).
\item Spawn positions: random poses on the table surface with physics
  settling.
\item Target assignment: selected to ensure the target is reachable
  (not fallen off table) after settling.
\end{itemize}

\noindent
The episode seed is computed as:
$\texttt{seed}_\text{ep} = \texttt{seed}_\text{base} + 1000 \times
\texttt{episode\_index}$.
All methods share identical initial conditions per episode
(same objects, same poses, same target).

\subsection{Lighting Modes}
\label{sec:supp_lighting}

\begin{figure*}[h]
\centering
\includegraphics[width=\linewidth]{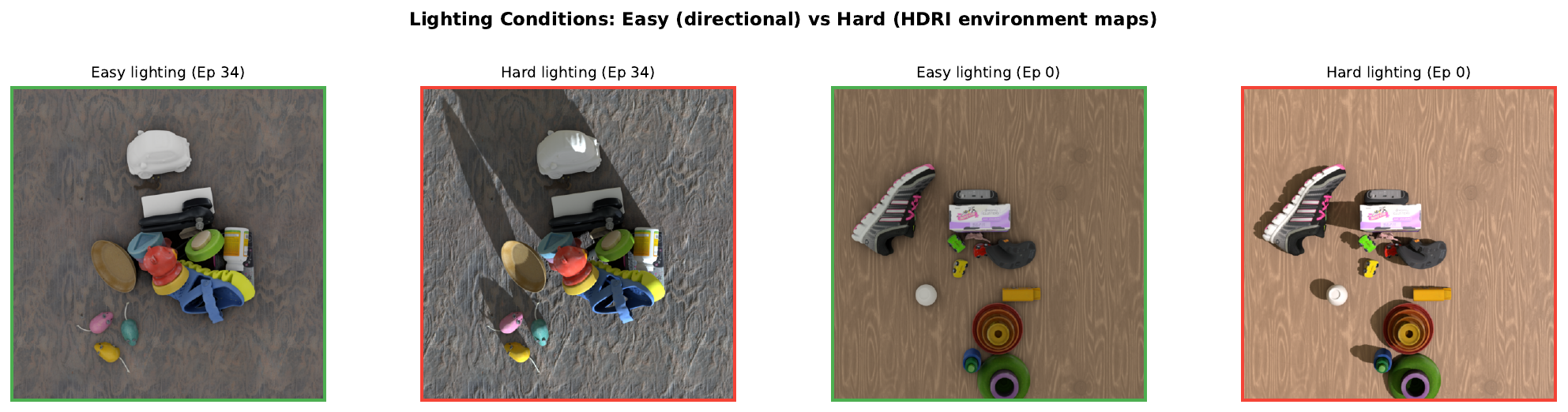}
\caption{Same scenes rendered under easy (directional light) and hard
(random HDRI environment map) lighting.  Hard lighting introduces strong
directional shadows, color casts, and specular highlights that
challenge perception modules.}
\label{fig:lighting}
\end{figure*}

\begin{itemize}
\item \textbf{Easy}: A fixed directional light matching PyBullet's
  default configuration---consistent, shadow-free, neutral color
  temperature.
\item \textbf{Hard}: A randomly-sampled indoor HDRI environment map
  per episode, producing variable shadow directions, color casts
  (warm/cool), and specular reflections on glossy objects
  (Figure~\ref{fig:lighting}).
\end{itemize}

\section{\red{Tool-Call Flow, Push--Grasp Selection, and Human Assessment}}
\label{sec:supp_toolflow}

\subsection{\red{Grasp vs.\ push: contextual selection}}
\label{sec:supp_grasp_vs_push}

\red{The near-balanced $49\%$/$47\%$ push--vs--grasp split after a successful
localization is not a fixed alternation: it reflects context-dependent selection
based on the state of the scene. To characterize these contexts, we tagged the
planner's free-form \texttt{reasoning} field in every push and grasp call issued
immediately after a \texttt{locate\_object} success, across 100 simulation
episodes (one seed, easy lighting; $n{=}196$ pushes, $n{=}188$ grasps).}

\begin{table}[h]
\centering
\small
\setlength{\tabcolsep}{3pt}
\begin{tabular}{lcc}
\toprule
Contextual cue in planner's reasoning & \texttt{push} & \texttt{grasp} \\
\midrule
Target/occluder \emph{hidden}             & $29.0\%$ & $30.8\%$ \\
Insufficient clearance around target  & $30.6\%$ & $17.4\%$ \\
Object needs \emph{reorientation} for stable grasp    & $7.9\%$  & $4.1\%$  \\
Object \emph{obstructs} the target directly            & $1.2\%$  & $5.7\%$  \\
Target object visible and reachable  & $5.5\%$  & $60.1\%$ \\
\bottomrule
\end{tabular}
\caption{\red{Fraction of push and grasp calls whose \texttt{reasoning} text
mentions each context cue (keyword-tagged; categories are not mutually
exclusive). Pushes are dominated by \emph{approach clearance} and
\emph{revealing hidden targets}; grasps are dominated by \emph{removing
a now-exposed obstructor or target}.}}
\label{tab:push_vs_grasp_contexts}
\end{table}

\red{Three patterns emerge. First, \textbf{pushes are preferred when the
target is not directly reachable}: $30.6\%$ of pushes cite clearing an
approach path or making room for the gripper, versus $17.4\%$ of grasps.
Second, \textbf{grasps dominate when an obstructor is already exposed
and lightweight enough to remove}: $60.1\%$ of grasps explicitly cite
completing the task or removing a known obstructor. Third,
\textbf{both primitives are used to reveal hidden objects} at similar
rates ($29$--$31\%$), which is what pushes the two totals so close
together in the aggregate: revealing an occluded region can be done
either by shoving a large item aside (push) or by lifting a small item
out (grasp), and the planner picks based on the item's manipulability.
Representative examples from the logs are given below.

\begin{quote}\small\itshape
Push: ``pushing the black boot to the left to reveal any objects hidden
underneath it.''

Push: ``the towel is still partially
covering the area. I will push it to the left to completely uncover any
hidden objects.''
\end{quote}
\begin{quote}\small\itshape
Grasp: ``removing the white bunny figurine to reveal any objects hidden
underneath it, as the target toy football is currently occluded.''

Grasp: ``the target fresh nectarine is now fully
exposed\ldots grasping it will complete the task.''
\end{quote}
}

When \texttt{locate\_object} returns \texttt{not\_found}, it retries in 87\% of cases rather than abandoning the target — the object is typically occluded rather than absent, so a retry after further decluttering is the productive response. After a push, it calls \texttt{locate\_object} in 78\% of cases: the displaced object is still present and its identity unchanged, so only its position needs updating. After grasping an occluder, the pattern inverts and the planner calls \texttt{survey\_scene} in 72\% of cases, since removing an object can expose previously hidden ones that no existing scene-graph entry describes.

\subsection{\red{Human Assessment of Action Decisions}}
\label{sec:supp_human}

\begin{table}[t]
\centering\footnotesize
\setlength{\tabcolsep}{6pt}
\renewcommand{\arraystretch}{1.15}
\begin{tabular}{lrrrr}
\toprule
 & \red{judgments} &
 \multicolumn{3}{c}{\red{rating (\%)}} \\
\cmidrule(lr){3-5}
\red{action} & \red{\% ($n$)} & \red{pos} & \red{neu} & \red{neg} \\
\midrule
\red{grasp} & \red{68.7 (206)} & \red{88.8} & \red{5.9} & \red{5.3} \\
\red{push} & \red{31.3 (94)} & \red{77.7} & \red{19.1} & \red{3.2} \\
\midrule
\red{\textbf{overall}} & \red{\textbf{100.0 (300)}} &
\red{\textbf{85.3}} & \red{\textbf{10.0}} & \red{\textbf{4.7}} \\
\bottomrule
\end{tabular}
\caption{\red{\textbf{Human assessment of \claro{}'s action decisions.} 300
judgments by eight independent annotators, each rating the effect of
\claro{}'s action on retrieving the target as positive, neutral, or
negative.}}
\label{tab:supp_human}
\end{table}

\red{
\paragraph{Protocol.} Each participant was assigned one randomly drawn trial per episode.
They were shown \claro{}'s earlier actions as context and the current action it had taken together with the reasoning it recorded, and rated that action's effect on retrieving the target as \emph{positive} (helps), \emph{neutral} (does not meaningfully help or hurt), or \emph{negative} (makes retrieval harder).
Whenever the rating was not positive, they also selected a reason: wrong object, wrong primitive, wrong direction, repeated failed action, or unnecessary action because the target was already reachable.
Because annotators see the action and its recorded reasoning but not its outcome, the rating evaluates the \emph{decision} rather than its execution.
We collected 300 completed judgments
from eight annotators, none of whom judged consecutive steps of the same rollout. These judgments cover every executed action in the real-robot episodes, so the grasp share ($68.7\%$) is higher than the post-localization split reported in \cref{sec:supp_grasp_vs_push}: a failed grasp is retried and counted again, and real scenes are sparser than the simulated ones, leaving the target directly reachable more often.

\paragraph{Ratings.} Table~\ref{tab:supp_human} summarizes the outcome. Across
all judgments 85.3\% of \claro{}'s decisions were rated positive, 10.0\%
neutral and 4.7\% negative. Grasps were rated positive in 88.8\% of judgments
and pushes in 77.7\%, but pushes drew the \emph{lower} negative rate, 3.2\%
against 5.3\%: what separates the two primitives is the neutral share, not the
negative one. Alongside the ambiguity over which direction
to push, a second case falls into the neutral category. \claro{} often clears
an object adjacent to the target whose removal eases access without being
strictly required: such an action neither fails to help nor makes retrieval
harder, and with only three options available an annotator records it as
neutral. The reason counts support this reading over the alternative that
neutral marks a wasted action: while 19.1\% of pushes were rated neutral, only
8.5\% were flagged as unnecessary because the target was already reachable, so
fewer than half of the neutral pushes were judged pointless.

}

\section{Supplementary Video}
\label{sec:supp_video}

The supplementary video shows some success and failure cases of the robot experiments with CLARO, compared to the baselines. In our robot setup, the robot arm might collide with an object which triggers a safety-terminate step on the robot controller. When this happens during an experiment, we manually reset the arm to a free space and resume the experiment from that step for all the baselines. We never manipulate the intermediate scene clutter.